\documentclass[letterpaper]{article} 
\usepackage{array}
\usepackage{tabularx}
\usepackage{xcolor}
\usepackage{colortbl}
\usepackage{makecell}
\usepackage{ragged2e}
\usepackage{threeparttable}
\usepackage{booktabs}
\usepackage{multirow}
\usepackage{amssymb}
\usepackage{aaai25}  
\usepackage{times}  
\usepackage{helvet}  
\usepackage{courier}  
\usepackage[hyphens]{url}  
\usepackage{graphicx} 
\usepackage{natbib}  
\usepackage{caption} 
\usepackage{algorithm}
\usepackage{algorithmic}
\usepackage{amsmath}

\usepackage{newfloat}
\usepackage{listings}
\DeclareCaptionStyle{ruled}{labelfont=normalfont,labelsep=colon,strut=off} 
\floatstyle{ruled}
\newfloat{listing}{tb}{lst}{}
\floatname{listing}{Listing}
\title{MoChat: Joints-Grouped Spatio-Temporal Grounding LLM for Multi-Turn Motion Comprehension and Description}
\author{
    Jiawei Mo\textsuperscript{\rm 1}, Yixuan Chen\textsuperscript{\rm 1}, 
    Rifen Lin\textsuperscript{\rm 1}, Yongkang Ni\textsuperscript{\rm 3}, 
    Min Zeng\textsuperscript{\rm 1},\\
    Xiping Hu\textsuperscript{\rm 2}\thanks{Co-corresponding author.}, 
    Min Li\textsuperscript{\rm 1}\thanks{Co-corresponding author.} 
}
\affiliations{
    \textsuperscript{\rm 1} School of Computer Science and Engineering, Central South University\\
    \textsuperscript{\rm 2} Shenzhen MSU-BIT University\\
    \textsuperscript{\rm 3} School of software, Xinjiang University\\
    mojiawei@csu.edu.cn, limin@mail.csu.edu.cn, huxp@smbu.edu.cn
}

\begin{document}

\maketitle

\begin{abstract}
Despite continuous advancements in deep learning for understanding human motion, 
existing models often struggle to accurately identify action timing and specific body parts, 
typically supporting only single-round interaction. 
Such limitations in capturing fine-grained motion details reduce their effectiveness in motion understanding tasks.
In this paper, we propose MoChat, a multimodal large language model capable of spatio-temporal grounding of human motion and understanding multi-turn dialogue context. 
To achieve these capabilities, we group the spatial information of each skeleton frame based on human anatomical structure and then apply them with Joints-Grouped Skeleton Encoder, 
whose outputs are combined with LLM embeddings to create spatio-aware and temporal-aware embeddings separately.
Additionally, we develop a pipeline for extracting timestamps from skeleton sequences based on textual annotations, 
and construct multi-turn dialogues for spatially grounding.
Finally, various task instructions are generated for jointly training. 
Experimental results demonstrate that MoChat achieves state-of-the-art performance across multiple metrics in motion understanding tasks, 
making it as the first model capable of fine-grained spatio-temporal grounding of human motion.
\end{abstract}
\begin{figure}[t]
    \centering
    \includegraphics[width=0.8\columnwidth]{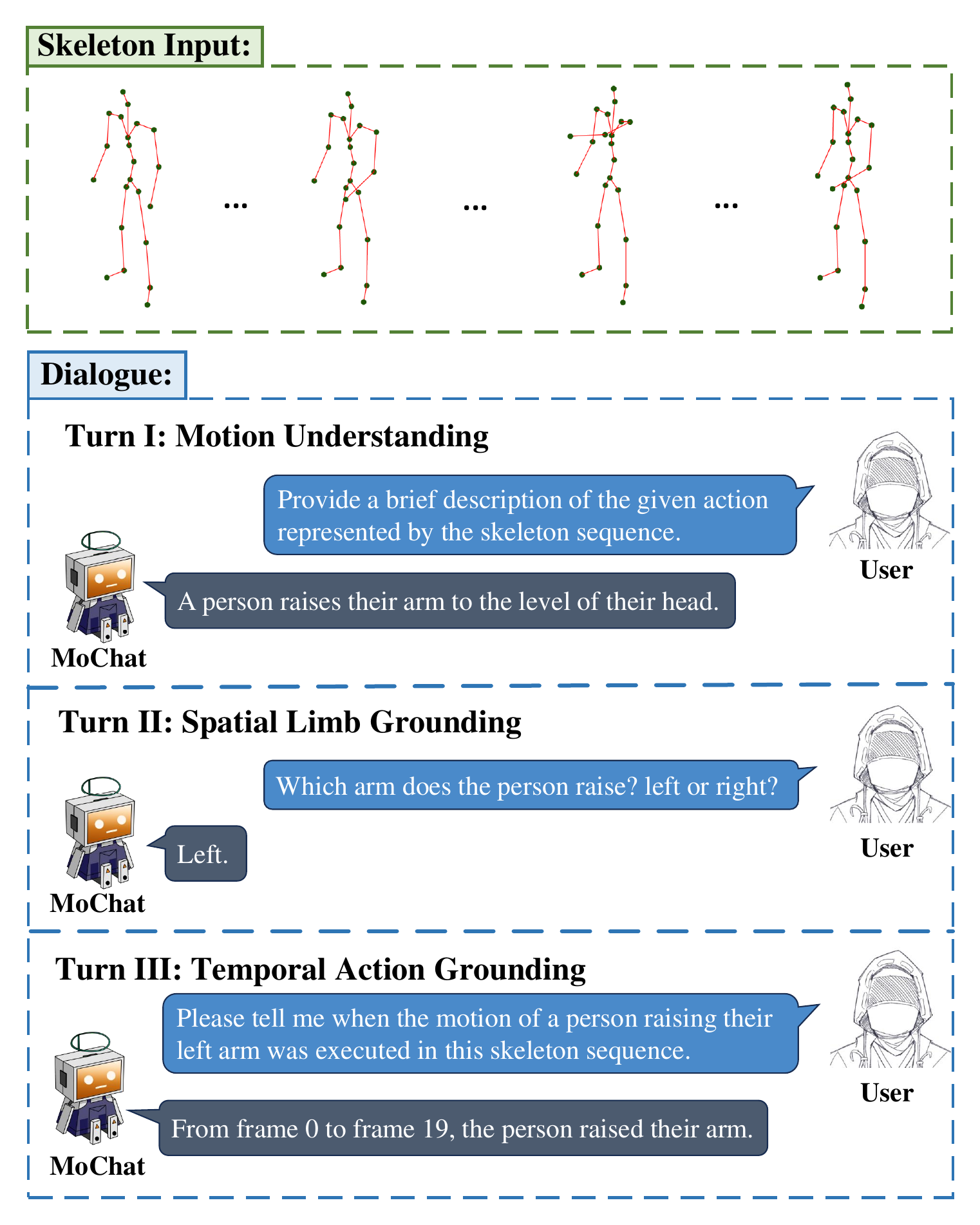} 
    \caption{Illustration of the multi-turn spatio-temporal grounding capabilities of MoChat. MoChat is a large language model designed for motion comprehension, with capabilities that extend beyond regular motion description. Specifically, MoChat can follow user instructions to summarize motion sequences (Turn I), pinpoint specific body parts involved in the motion (Turn II), and ground the start and end frames corresponding to user queries (Turn III).}
    \label{intro}
\end{figure}
\section{Introduction}
The analysis and understanding of human motion have extensive applications across multiple fields, including human-computer interaction, virtual reality, security surveillance, medical rehabilitation, and sports broadcasting.
Recent breakthrough of multimodal large language models (MLLMs), such as Flamingo \cite{Flamingo}, GPT-4V \cite{openai2024gpt4technicalreport} and CogVLM \cite{Hong2024CVPR}, has enabled AI to achieve open-vocabulary human motion understanding.
Existing works on MLLM-based human motion understanding can be broadly classified into two categories:
the first category encompasses models focused on RGB image and video understanding, such as VideoChat \cite{2023videochat} and BLIP-2 \cite{blip2}, which are not specifically tailored for human motion understanding tasks;
the second category comprises specialized models designed explicitly to interpret human motion from motion capture data, showcasing advanced performance in analyzing motion, exemplified by TM2T \cite{tm2t} and MotionGPT \cite{jiang2024motiongpt}.
However, these models still struggle to accurately ground specific time periods and body parts involved in motion, 
which limits their performance in motion understanding tasks.

The challenge of building such motion understanding models lies in accurately modeling the relationships between motion sequences and captions, and incorporating the temporal dimensions essential for understanding motion.
For the first challenge, recent research \cite{zhu2024languagebind} has demonstrated the efficacy of pre-trained large language models (LLMs) in modeling relationships between diverse non-textual modalities and textual data. 
Specifically, motion sequences can be regarded as a unique form of language. 
By utilizing an projector, these sequences can be fine-tuned to facilitate the conversion of motion information into descriptive text.
Additionally, in the action recognition field, studies \cite{yan2023skeletonmae,Huang_Huang_Ouyang_Wang_2020} have shown that grouping keypoints can enhance the representation of action features.
For the second challenge, existing video captioning models \cite{timechat,momentor} are capable of extracting the time intervals in videos that correspond to specific captions. 
Therefore, it is promising to train a model capable of locating the spatial and temporal positions of specific action sequences. 

In this work, we propose MoChat, a multimodal large language model that is capable of spatio-temporal grounding in human motion understanding, facilitated by multi-turn dialogue context.
To enable the model's understanding of motion sequences, we first pre-train a Transformer-based \cite{transformer} skeleton encoder. 
The keypoints are partitioned into four groups based on the human anatomical structure for motion encoding, enhancing the encoder's geometric perception. 
The resulting motion features are then converted through a lightweight projector into LLM-compatible tokens, which are subsequently combined with text instruction tokens as input into the LLM.
This allows the model to comprehend the semantics of the motion sequence and generate descriptive text for the motion sequence.
Meanwhile, by calculating the similarity between the LLM's hidden states and the motion tokens, the temporal boundaries corresponding to the text are regressed. 
Additionally, to construct dialogue data for training, we develop a pipeline for extracting timestamps from the motion caption datasets, and create multi-turn spatial dialogues by keyword matching. 
Using the resulting multi-task instruction set, we conduct a two-stage joint training of MoChat, which enhances its detailed action understanding capabilities in both temporal and spatial dimensions.
We validate our model through extensive experiments on the HumanML3D dataset \cite{humanml3d}, covering the tasks of Motion Understanding, Spatial Limb Grounding, and Temporal Action Grounding, evaluated using tranditional metrics and GPT-4.
The results demonstrate that MoChat achieves state-of-the-art performance, highlighting its fine-grained spatio-temporal motion understanding capabilities.
Our contributions can be summarized as follows:

\begin{itemize}
    \item We propose MoChat, a motion understanding multimodal large language model that comprehends motion sequences, 
    accurately captions the movement of specific body parts, 
    and precisely identifies the time boundaries corresponding to user instructions. 
    To the best of our knowledge, MoChat is the first MLLM capable of spatio-temporal grounding of actions in skeleton sequences.
    \item We develop a semi-automated pipeline to extract timestamps from the motion caption datasets, and construct multi-turn spatial dialogues, both of which are used to create a multi-task instruction set for joint training.
    \item Comprehensive experiments validate the advanced motion understanding capabilities of MoChat, demonstrating its spatial and temporal grounding abilities. 
    Our model introduces functionalities not found in existing motion understanding models, making it more versatile and user-friendly.
\end{itemize}

\section{Related Work}
\subsubsection{Motion Understanding Models}
Motion understanding tasks can generally be categorized into fixed-class action recognition, which involves a predefined set of classes, 
and open-vocabulary motion understanding, which does not restrict the number of classes. 
In the branch of fixed-class action recognition, numerous skeleton-based methods have been proposed. \cite{2sagcn,poseconv,ctrgcn} 
For instance, ST-GCN \cite{stgcn} applies 3D graph convolution to human skeleton sequences across both temporal and spatial dimensions to extract action features. 
With the rise of self-supervised learning and Transformers \cite{transformer}, there has been a shift towards exploring Transformer-based self-supervised action recognition. \cite{crossclr,hitrs} 
One such method is GL-Transformer \cite{gltrans}, which constructs pretext tasks for amplitude and displacement recovery using the relative and absolute positions of joints, 
enabling effective representation of skeleton sequences without reliance on action labels. 

With the advancement of LLMs, open-vocabulary motion understanding tasks have become feasible. 
The models typically involve a motion encoder combined with a language model to comprehend motion sequences. 
A notable example is TM2T \cite{tm2t}, which employs VQVAE \cite{vqvae} to obtain discrete motion tokens from a codebook. 
These motion tokens and their corresponding text tokens are then fed into simple neural machine translators (NMT) for both motion-to-text and text-to-motion conversion, enabling bidirectional matching.
MotionGPT \cite{jiang2024motiongpt} and AvatarGPT \cite{AvatarGPT} replace NMT with LLMs equipped with projector, fine-tuned with instructions to enable understanding and generation of motion sequences under various conditions.
However, these methods have not fully exploited the comprehension capabilities of LLMs, primarily due to insufficient training instructions and the limited representational power of the encoders.
\begin{figure*}[t]
    \centering
    \includegraphics[width=1.0\textwidth]{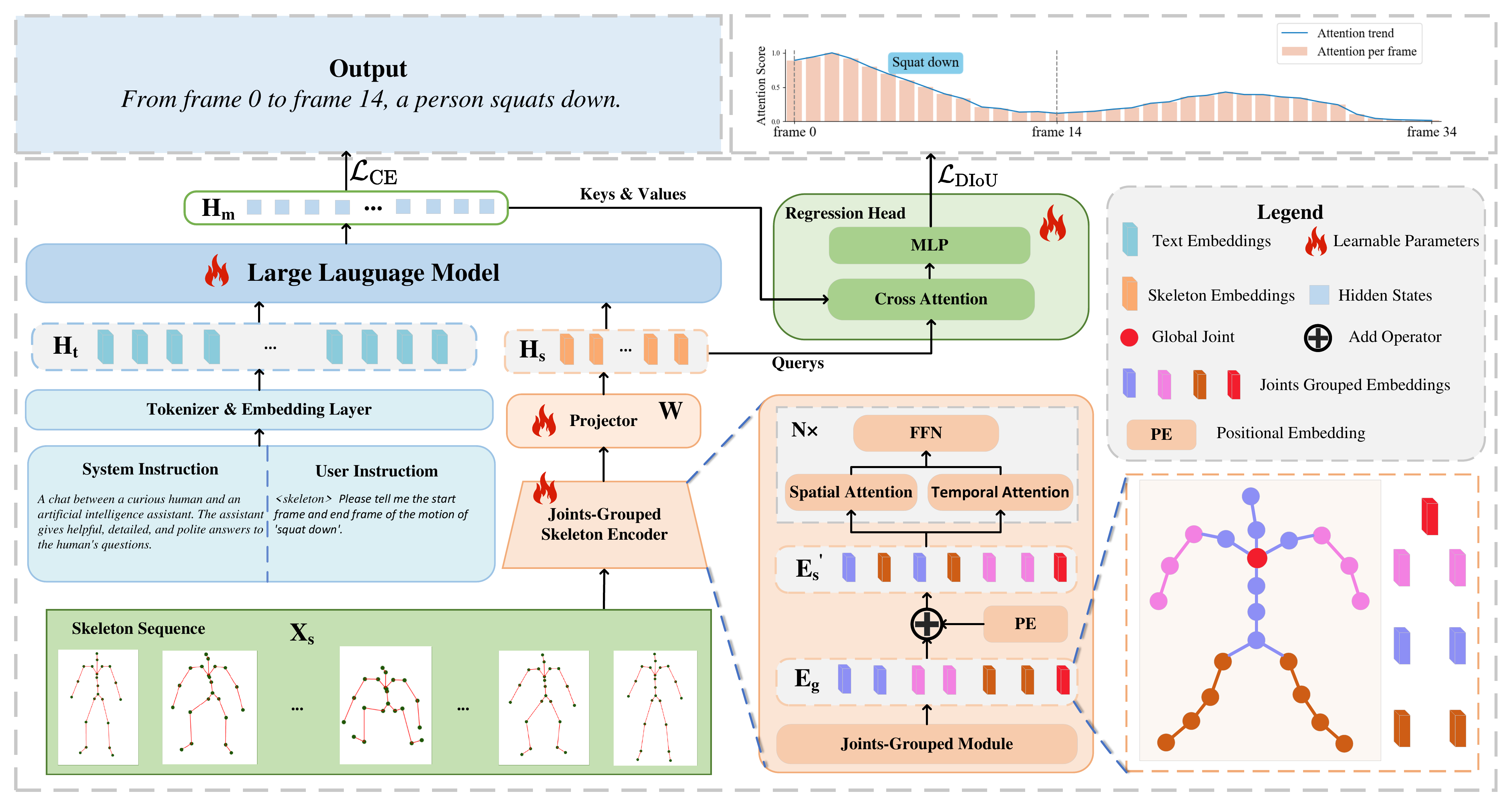} 
    \caption{Overview of MoChat. Given a skeleton motion sequence as input, 
    \textbf{(a) Joints-Grouped Skeleton Encoder} first extracts motion features by grouping and embedding the joints separately. 
    Then, \textbf{(b) Projector} converts these features into motion tokens $H_s$ in the language latent space. 
    These motion tokens $H_s$ are concatenated with instruction tokens $H_t$ and input to a \textbf{(c) Large Language Model (LLM)}. 
    The LLM's final hidden states $H_m$ are decoded into appropriate responses and passed to a \textbf{(d) Regression Head} to obtain the corresponding timestamps.}
    \label{overview}
\end{figure*}
\subsubsection{Vision-Language Models}
The development of large language models (LLMs) has significantly advanced the field of vision-language models, with notable progress in both image-language models \cite{openai2024gpt4technicalreport,llava1.5} and video-language models \cite{chatunivi,timechat}.
In the domain of image-language models, LLaVA-1.5 \cite{llava1.5} employs VIT \cite{CLIP} as the image encoder and Vicuna \cite{vicuna2023} as the language decoder. 
A lightweight projector is used to map image embeddings into the language latent space, enabling LLMs to understand visual content. 
In contrast, CogVLM \cite{wang2024cogvlmvisualexpertpretrained} introduces a visual expert module that is equivalent in size to the LLM. 
Yet this approach doubles the inference parameters of the LLM, which presents challenges during deployment. 
BLIP-2 \cite{blip2} pre-trains a BERT-based \cite{devlin2018bert} Q-Former to align visual and textual information, using a fixed-length learnable query vector to extract semantic information from images. 
However, this approach overly compresses the information, limiting the model's ability to capture intricate image details.
For video understanding, ChatUnivi follows the LLaVA's projector approach, also compressing information by aggregating dynamic visual tokens across different frames. 
On the other hand, TimeChat adopts the InstructBLIP \cite{dai2023instructblip} strategy to encode temporal information through textual instructions. 
Besides, it employs a sliding window to segment video frames, encoding them with multiple Q-Formers. 
These approaches enhance TimeChat’s temporal awareness but it struggles with continuous temporal concept comprehension.
Additionally, previous work \cite{TellingLeft} has revealed significant challenges in vision models' handling of “geometry-aware” semantic correspondences. 
For example, these models often misinterpret spatial relationships, such as confusing the left and right sides of the image with the left and right sides of the objects within it, 
which hampers their spatial grounding capabilities.
To address these limitations, we propose MoChat, the first motion understanding model that achieves accurate spatio-temporal grounding. 

\section{MoChat: A Chat MLLM for Motion}
In this section, we introduce MoChat, a multimodal large language model capable of spatio-temporal grounding in human motion understanding, facilitated by multi-turn dialogue context. 
The inclusion of two novel modules, the Joints-Grouped Skeleton Encoder and the Regression Head, enhances MoChat's ability to finely understand motions and accurately ground the start and end frames of instruction-corresponding motions. 
To further empower MoChat to follow human instructions and understand context in complex multi-turn, multi-task dialogues, we construct such dialogues for spatial fine-grained motion understanding and develop a pipeline for timestamp extraction. 
Based on these dialogues, we perform a two-stage integrated instruction tuning on a pre-trained LLM to create MoChat.

\subsection{Overall Framework}
As illustrated in Fig. \ref{overview}, MoChat is composed of a spatio-aware Joints-Grouped Skeleton Encoder (JGSE), a LLM equipped with projector, and a regression head.
Given an input skeleton sequence with $T$ frames, $X_{s} = \left\{ X_{s}^{i} \right\}_{i=1}^{T}$, the skeleton encoder JGSE first extracts motion features while maintaining the same sequence length. 
Then, a projector converts these features into motion tokens $H_s$, which are mapped to the language latent space. 
These motion tokens $H_s$ are concatenated with input instruction tokens $H_t$ and fed into a Large Language Model (LLM). 
The LLM's final hidden states $H_m$ are then decoded into appropriate responses, which are passed to a regression head to obtain the corresponding timestamps simultaneously.
\subsubsection{Joints-Grouped Skeleton Encoder}
Previous transformer based models typically apply positional encoding to skeleton joints based on the specific order determined by the joint numbering scheme. 
However, different skeleton types have different joint numbering orders, which forces models to undergo retraining when the skeleton type changes. 
While this approach is effective for handling specific skeleton types, it ultimately limits the model's ability to generalize and effectively represent other skeleton types.
In transformers, positional embeddings are initially designed to reinforce the positional relationships within a sequence, making the order of the input sequence critically important.
This implies that when a frame of skeleton joints is used as the input sequence, different orders of the joints can significantly alter the transformer's encoding output.

With this consideration in mind, we choose GL-Transformer and modified its position encoding method and embedding strategy to develop a new model, the Joints-Grouped Skeleton Encoder.
For each skeleton frame, which includes \( M \) joints denoted as \( X_{s}^{i} = \{ j_{k} \}_{k=1}^{M} \), we partition the skeleton joints \( j_{k} \) into four groups \( G_g \), based on human anatomical structure, where:
\begin{equation}
    g \in \left\{
    \begin{aligned}
        &\text{Arm (A)},  &\text{Leg (L)}, \\
        &\text{Trunk (T)}, &\text{Global Joint (GJ)}
    \end{aligned}
    \right\}.
\end{equation}
The Global Joint (GJ) is derived by applying a weighted combination of all joints, and it is used to capture the holistic representation of the skeleton.

Each group of joints is then embedded, resulting in embeddings \( E_{\text{A}} \), \( E_{\text{L}} \), \( E_{\text{T}} \), and \( E_{\text{GJ}} \) for the Arm, Leg, Trunk, and Global Joint groups, respectively. 
These embeddings are subsequently concatenated to form the final skeleton embedding:
\begin{equation}
    E_g = \text{Concat}(E_{\text{A}}, E_{\text{L}}, E_{\text{T}}, E_{\text{GJ}}).
\end{equation}

Next, we successively add spatial and temporal positional embeddings to the ordered skeleton embedding sequence \(E_s\) to reinforce both spatial and temporal positional representation. 
To facilitate the exchange of information aggregated to the joints, \(E_s\) is then restored to \(E_s'\) according to the original joint numbering order and passed to the \(N\)-layer transformer encoder.

\subsubsection{Language Module}
We follow the LLaVA-1.5 \cite{llava1.5} approach to construct the language module, which is based on the large language model Vicuna \cite{vicuna2023} equipped with a linear projector. 
After being processed by the JGSE, the motion features are converted into motion embedding tokens \(H_s\) through a trainable projection matrix \(W\). 
This projection maps the motion features into the language embedding space while preserving the sequence length \(T\), resulting in motion embedding tokens \(H_s\).

To enhance the LLM's ability to follow user instructions, we design a prefix system instruction to make the model more user-friendly. 
The user input is referred to as the alternative user instruction, and the \texttt{<skeleton>} placeholder indicates the position of the skeleton sequence. 
After concatenating the system and user instructions, the instruction embedding tokens \(H_t\) are generated by the LLM's tokenizer and embedding layer. 
Finally, the motion embedding tokens \(H_s\) are inserted into the instruction embedding tokens \(H_t\) at the placeholder position, and the combined sequence is fed into the LLM.

The output of the LLM, specifically its final hidden states \(H_m\), is then processed to generate the model's predictions. 
These final hidden states \(H_m\) are passed through a linear layer to produce the logits \(\mathbf{z}\), which are subsequently decoded into the output \(X_o\). 
At training time, the cross-entropy loss is calculated between the logits \(\mathbf{z}\) and the labels \(X_{gt}^{\text{id}}\) 
(the token IDs corresponding to the ground truth \(X_{gt}\), which is obtained by shifting the dialogue \(X_t\) one position to the left), 
while the inserted skeleton sequence does not contribute to the loss calculation:
\begin{equation}
    \mathcal{L}_{\text{CE}} = -\sum_{i} X_{gt}^{\text{id}(i)} \log \sigma(\mathbf{z}^{(i)}),
\end{equation}
where \(\sigma(\cdot)\) denotes the softmax function applied to the logits \(\mathbf{z}\).

\subsubsection{Regression Head}
For precisely grounding the time boundaries, we design a regression head, which is responsible for predicting the start frame \(\text{ID}_{\text{start}}\) and the end frame \(\text{ID}_{\text{end}}\). 
To compute the start and end frame IDs corresponding to the language, we naturally consider calculating the similarity between the motion embedding tokens \(H_s\) and the LLM hidden states \(H_m\). 
In this process, the motion embedding tokens \(H_s\) are fed into the regression head as \(Queries\), while the LLM hidden states \(H_m\) serve as \(Keys\) and \(Values\). 
We employ the scaled dot-product attention mechanism to compute the attention weights:
\begin{equation}
    \text{W}_{\text{cross}} = \text{softmax}\left(\frac{QK^T}{\sqrt{d_k}}\right),
\end{equation}
where \(Q\) represents the queries, \(K\) represents the keys, and \(d_k\) is the dimension of the keys. 
The resulting attention weights \(\text{W}_{\text{cross}} \in \mathbb{R}^{T \times N}\). 
We then focus on the weight of the \texttt{[BOS]} token \(\text{W}_0 \in \mathbb{R}^{T \times 1}\), as it is the most significant token for representing the entire sequence.

Subsequently, a Multi-Layer Perceptron (MLP) is used to regress the start and end frame IDs:
\begin{equation}
    \text{IDs} = \text{MLP}(W_0^T \cdot H_s),
\end{equation}
where \(H_s \in \mathbb{R}^{T \times D}\), with \(D\) being the hidden dimension of the LLM.
The output \(\text{IDs}\) corresponds to \([\text{ID}_{\text{start}}, \text{ID}_{\text{end}}]\).

Then, for stable convergence, the DIoU loss \cite{zheng2020diou} between the predicted and ground truth IDs is calculated as:
\begin{equation}
    \mathcal{L}_{\text{DIoU}} = 1 - \left( \text{IoU} - \frac{d^2(\text{ID}_{\text{start}}, \text{ID}_{\text{end}}, \text{ID}_{\text{start}}^{\text{gt}}, \text{ID}_{\text{end}}^{\text{gt}})}{c^2(\text{ID}_{\text{start}}, \text{ID}_{\text{end}}, \text{ID}_{\text{start}}^{\text{gt}}, \text{ID}_{\text{end}}^{\text{gt}})} \right),
\end{equation}
where \(\text{IoU}\) denotes the Intersection over Union. 
The \(d^2(\cdot)\) term represents the squared Euclidean distance between the center points of the predicted and ground truth intervals, 
while the \(c^2(\cdot)\) term normalizes this distance by the square of the length of the union interval.

The final loss is a combination of both:
\begin{equation}
    \mathcal{L} = \mathcal{L}_{\text{CE}} + \lambda_{\text{DIoU}} \mathcal{L}_{\text{DIoU}},
\end{equation}
where $\lambda_{\text{DIoU}}$ is a hyperparameter that balances the two losses.

\subsection{Data Construction}
We construct motion understanding dialogues using the motion caption dataset. 
We initially design instructions such as \textit{Provide a brief description of the given action represented by the skeleton sequence} and directly use the corresponding motion caption as the answer for constructing basic motion understanding dialogues.

\begin{threeparttable}[htbp]
    \begin{tabularx}{\linewidth}{>{\centering\arraybackslash}m{2cm}|>{\RaggedRight\arraybackslash}m{5.5cm}}
        \toprule
        \rowcolor{gray!10} \multicolumn{2}{l}{\textbf{Temporal Grounding Dialogues}} \\
        \midrule
        \textbf{Dialogue Templates} & 
        Please tell me when \texttt{<motion>} was executed in this skeleton sequence. \newline
        From \texttt{<frameid\_1>} to \texttt{<frameid\_2>}, \texttt{<motion>}\\
        \midrule
        \textbf{Example} & 
        \textbf{Q:} Please tell me when \textbf{\textit{A person bends forward}} was executed in this skeleton sequence. \newline
        \textbf{A:} From \textbf{\textit{frame 20}} to \textbf{\textit{frame 33}}. \textbf{\textit{A person bends forward}}. \\
        \midrule
        \rowcolor{gray!10} \multicolumn{2}{l}{\textbf{Spatial Gap-filling Dialogues}} \\
        \midrule
        \textbf{Instruction Templates} & 
        \texttt{<motion\_with\_gap>}, Complete the content in brackets with \textbf{\textit{left} or \textit{right}}. \\
        \midrule
        \textbf{Example} & 
        \textbf{Q:} Person leans forward goes onto knees whilst first putting \textbf{(\rule{5mm}{0.4pt})} hand on ground for support and stays on knees. Complete the content in brackets with left or right. \newline
        \textbf{A:} \textbf{\textit{Left}}. \\
        \midrule
        \rowcolor{gray!10} \multicolumn{2}{l}{\textbf{Spatial Multi-turn Dialogues}} \\
        \midrule
        \textbf{Instruction Templates} & 
        What actions is the person's \texttt{<body\_part>} performing?  \newline
        Tell me about the actions performed by the person's \texttt{<body\_part>}. \\
        \midrule
        \textbf{Example} & 
        \textbf{Q:} Tell me about the actions performed by the person's \textbf{\textit{torso}}. \newline
        \textbf{A:} The person walked backwards slowly, their \textbf{\textit{torso}} remaining upright, before stepping forward with a forceful kick. \newline
        \textbf{Q:} What actions is the person's \textbf{\textit{arm}} performing? \newline
        \textbf{A:} A person bends their left arm at the elbow and raises their right \textbf{\textit{arm}} towards it, then lowers both arms. \\
        \bottomrule        
    \end{tabularx}
    \caption{Dialogue Examples. \textit{Q} represents the human instruction, and \textit{A} represents the ground truth answer. Only a subset of the templates is shown here; the complete set can be found in the supplementary material.}
    \label{instruction_templates}
\end{threeparttable}

\subsubsection{Spatial Dialogues Construction}
We construct multi-turn dialogues for spatial fine-grained motion using keyword matching.
First, we select keywords such as \textit{foot}, \textit{leg}, \textit{hand}, \textit{arm} and \textit{torso} based on human anatomical structure. 
Next, we create instruction templates, as shown in Tab. \ref{instruction_templates}, where the \texttt{<body\_part>} placeholder in the instruction can be replaced with these keywords. 
Captions containing the corresponding keywords are then selected as responses.
If a caption involves multiple body parts, it is split into separate turns, with each turn's response describing the motion of a single body part. 
For spatial relationships, we design gap-filling dialogues based on captions that include spatial keywords such as \textit{left} and \textit{right}.
Specifically, we ensure a balanced distribution of different answers to prevent model bias.

\subsubsection{Timestamps Extraction Pipeline}
We develop a pipeline for extracting timestamps from skeleton sequences based on textual annotations.
To avoid any potential bias in subsequent GPT-4 scoring,
GLM-4 \cite{glm2024chatglm} is employed, with the instruction shown in the supplementary material,
 to determine the atomic action referenced in the captions
and to identify one corresponding joint and axis (X for left-right, Y for height, Z for front-back) exhibiting the most significant variation. 
This process simplifies the task of accurately assigning timestamps to each individual action.
The selection of joints and axes is further refined based on motion data. 
Following the analysis from GLM-4, the selected motion data is first smooth-filtered. 
Subsequently, extreme points and the differences between them are computed, 
allowing for the identification of the start and end frame IDs that correspond to the atomic action with the maximum variation.
After extraction, a manual review is conducted, and the results are used to construct the temporal grounding dialogues as shown in Tab. \ref{instruction_templates}.


\subsection{Training Strategy}
Our training strategy consists of three stages: pre-training the skeleton encoder, aligning motion-language embeddings, and fine-tuning the model end-to-end.
In the latter two stages, we conduct an integrated instruction tuning process on a pre-trained LLM, which involves two sequential steps while keeping the JGSE frozen.

For the skeleton encoder pre-training, we unsupervisedly train the JGSE on skeleton sequences, following the data preprocessing and pretext tasks outlined in \cite{gltrans}.

Next, we jointly train the projector and regression head, with the LLM frozen, using multi-task instruction set to align the motion embeddings with the LLM embeddings. 
Specifically, we merge the dialogues constructed in the previous subsubsection and randomly sample a batch for each iteration.
The human instructions from these dialogues and motion sequences serve as loss-irrelevant inputs to the LLM, while only the dialogue responses are used as loss-relevant inputs.
We then conduct autoregressive training to generate the next token for the input dialogues and motion sequences, 
extracting timestamps from the ground truth responses to calculate the DIoU loss. 
Finally, we fully fine-tune the entire LLM and projector using the same instruction set for further improvement.

\setlength{\tabcolsep}{3pt}
\begin{table*}[htbp]
    \centering
    \begin{tabular}{lcccccc}
    \toprule
    \textbf{Methods} & \textbf{BLEU@1 ↑} & \textbf{BLEU@4 ↑} & \textbf{ROUGE ↑} & \textbf{CIDEr ↑} & \textbf{BERTScore ↑} & \textbf{GPT4Score ↑} \\
    \midrule
    TM2T \cite{tm2t}          & 48.90 & 7.00 & 38.10 & 16.80 & 32.20 & -- \\
    MotionGPT \cite{jiang2024motiongpt}    & 48.20 & 12.47 & 37.40 & 29.20 & 32.40 & 5.14 \\
    AvatarGPT \cite{AvatarGPT}   & 49.28 & 12.70 & 40.44 & 32.65 & \textbf{53.58} & -- \\
    Baseline   & 59.81 & 19.26 & 45.86 & 45.09 & 43.57 & 5.21 \\
    MoChat (Ours)   & \textbf{61.75} & \textbf{21.60} & \textbf{47.59} & \textbf{51.57} & \underline{45.59} & \textbf{5.99} \\
    MoChat-R (Ours) & \underline{60.06} & \underline{21.30} & \underline{46.08} & \underline{46.57} & 42.56 & 5.25 \\
    \bottomrule
    \end{tabular}
    \caption{Comparison of Motion Understanding task on HumanML3D dataset. 
    MoChat-R refers to MoChat with a regression head. 
    The $\uparrow$ symbol indicates that a higher value is better. 
    Bold and underline indicate the best and the second best result.}
    \label{motionUnderstanding}
\end{table*}

\setlength{\tabcolsep}{1pt}
\begin{table*}[htbp]
    \centering
    \begin{tabular}{l|c|c|cccccc}
    \toprule
    \textbf{Models} & \textbf{Modules} & \textbf{Instruction Sets} & \textbf{BLEU@1 ↑} & \textbf{BLEU@4 ↑} & \textbf{ROUGE ↑} & \textbf{CIDEr ↑} & \textbf{BERTScore ↑} & \textbf{GPT4Score ↑} \\
    \midrule
    Baseline & GLTE+Vicuna & & 59.85 & 20.80  & 45.46 & 44.88 & 41.63 & 4.74 \\
    MoChat & JGSE+Vicuna &BMUD & 61.36 & 21.30  & 46.69 & 47.98 & 44.14 & 5.62 \\
    MoChat-R & JGSE+Vicuna+RH &  & 60.11 & 20.34 & 45.86 & 46.45 & 42.84 & 5.10 \\
    \midrule
    Baseline & GLTE+Vicuna & & 59.95 & 20.51 & \textbf{47.64} & 49.30  & 44.28 & 5.40 \\
    MoChat & JGSE+Vicuna &BMUD+SD & 60.81 & 20.87 & 47.04 & 50.60  & 44.60  & 5.96 \\
    MoChat-R & JGSE+Vicuna+RH &  & 60.31 & 20.64 & 45.87 & 46.65 & 42.84 & 5.19 \\
    \midrule
    Baseline & GLTE+Vicuna & & 59.81 & 19.26 & 45.86 & 45.09 & 43.57 & 5.21 \\
    MoChat & JGSE+Vicuna &BMUD+SD+TGD & \textbf{61.75} & \textbf{21.60}  & 47.59 & \textbf{51.57} & \textbf{45.59} & \textbf{5.99} \\
    MoChat-R & JGSE+Vicuna+RH &  & 60.06 & 21.30  & 46.08 & 46.57 & 42.56 & 5.25 \\
    \bottomrule
    \end{tabular}
    \caption{Ablation study on the Motion Understanding task across different models and instruction sets. 
    The module names GLTE, JGSE, and RH refer to Global-Local Transformer Encoder, Joints-Grouped Skeleton Encoder, and Regression Head, respectively. 
    BMUD+SD+TGD indicates that the model was jointly trained on Basic Motion Understanding Dialogue, Spatial Dialogue, and Temporal Grounding Dialogue. 
    The $\uparrow$ symbol indicates that a higher value is better. Bold indicates the best result.}
    \label{ablationMU}
\end{table*}

\section{Experiments}
\subsection{Datasets and Evaluation Metrics}
\subsubsection{HumanML3D}
The HumanML3D dataset \cite{humanml3d} is used for training and evaluation, containing 14,616 motion sequences and 44,970 motion captions. 
The dataset is divided into training, validation, and test sets, with 80\%, 5\%, and 15\% of the data allocated to each set, respectively. 
We utilize 22-joint SMPL \cite{SMPL:2015} skeleton sequences and construct the multi-task dialogues from its training and test sets.

\subsubsection{Evaluation Metrics}
We evaluate our model on three tasks: Motion Understanding, Spatial Limb Grounding, and Temporal Action Grounding.
For the Motion Understanding task, we follow the approach in \cite{tm2t}, utilizing linguistic metrics including BLEU \cite{papineni2002bleu}, ROUGE \cite{lin2004rouge}, CIDEr \cite{vedantam2015cider}, and BERTScore \cite{BERTScore}. 
Additionally, as pointed out by \cite{zheng2023judging}, GPT-4 can be used to judge the results generated by LLMs. 
Therefore, we construct a prompt containing the reference captions and the outputs from all evaluated models for each test sample. 
GPT-4 is then required to assign a score between 0 and 10 based on the similarity between the model outputs and the reference captions. 
The average of these scores is computed to obtain the GPT4Score.
For the Spatial Limb Grounding task, we use accuracy as the evaluation metric, as the spatial test set is based on gap-filling dialogues.
For the Temporal Action Grounding task, the evaluation metric is ``R@1, IoU = $\mu$,'' which denotes the percentage of retrieved frame IDs with an intersection over union (IoU) greater than $\mu$ compared to the ground truth.


\subsection{Implement Details}
We adopt the pre-trained Vicuna-v1.5-13B model \cite{vicuna2023} as the language foundation model. 
All models are trained on 8 $\times$ Nvidia A800 GPUs. The \(\lambda_{\text{DIoU}}\) is set to 5.
Detailed training configurations and hyperparameters are provided in the supplementary material.

\subsection{Comparisons with State-Of-The-Art Methods}
We evaluate MoChat with state-of-the-art methods on three task including Motion Understanding, Spatial Limb Grounding and Temporal Action Grounding.
We use an unmodified GL-Transformer \cite{gltrans} as the skeleton encoder for the baseline model, with the LLM component kept consistent across all models.
The model that includes both the Joints-Grouped Skeleton Encoder and the Regression Head is referred to as MoChat-R, 
while the model without the Regression Head is referred to as MoChat.

\subsubsection{Comparisons on Motion Understanding}
The Motion Understanding task involves generating a brief caption based on a given motion sequence. 
We directly adopt the linguistic results from AvatarGPT \cite{AvatarGPT} and use the suggested evaluation method to assess MoChat. 
For a fair comparison, we evaluate MotionGPT using the motion data as described in its paper, 
and the resulting captions are evaluated by GPT-4. 
As shown in Tab. \ref{motionUnderstanding}, MoChat significantly outperforms recent works on the Motion Understanding task.

\subsubsection{Comparisons on Spatial Limb Grounding}
The Spatial Limb Grounding task involves identifying which body part is responsible for the action in a given motion sequence. 
Following the data processing methods outlined in previous sections, we constructed 2,574 gap-filling questions from the HumanML3D test set to evaluate the model. 
Since current motion understanding models lack spatial grounding capabilities, we opted to use the multimodal model GPT-4V for evaluation. 
The motion sequences were rendered into human motion videos, from which 10 frames were evenly sampled. 
These 10 images were then used to assess GPT-4V's spatial limb grounding capability via API calls.
As shown in Tab. \ref{LR1}, MoChat achieves the highest accuracy of 85.70\%, demonstrating its strong capability in spatial limb grounding.
\setlength{\tabcolsep}{10pt}
\begin{table}[ht]
    \centering
    \begin{tabular}{lc}
    \toprule
    \textbf{Model} & \textbf{Acc. ↑} \\
    \midrule
    GPT-4V             & 68.02 \\
    Baseline        & 80.12 \\
    MoChat (Ours) & \textbf{85.70} \\
    MoChat-R (Ours)     & 81.90 \\
    \bottomrule
    \end{tabular}
    \caption{Comparison of Spatial Limb Grounding task on spatial test dataset. 
    MoChat-R refers to MoChat with a regression head. 
    The $\uparrow$ symbol indicates that a higher value is better. 
    Bold and underline indicate the best and the second best result.
    }
    \label{LR1}
\end{table}

\subsubsection{Comparisons on Temporal Action Grounding}
The Temporal Action Grounding task requires the model to accurately locate the time range corresponding to user instructions. 
Since current motion understanding models lack temporal grounding capabilities, we opted to evaluate the time-sensitive video understanding model TimeChat. 
Specifically, we construct a test set containing 233 samples to assess models' performance. 
As shown in Tab. \ref{time}, although MoChat-R slightly underperformed MoChat in the previous two tasks, it surpassed other models in the Temporal Action Grounding task.

\setlength{\tabcolsep}{4pt}
\begin{table}[ht]
    \centering
    \begin{tabular}{lcc}
    \toprule
    \textbf{Model} & \textbf{R@1 (IoU=0.5) ↑} & \textbf{R@1 (IoU=0.7) ↑} \\
    \midrule
    TimeChat          & 2.10  & 0.40 \\
    Baseline       & 12.45 & \underline{6.87} \\
    MoChat (Ours)       & \underline{19.31} & 5.58 \\
    MoChat-R (Ours) & \textbf{21.89} & \textbf{12.02} \\
    \bottomrule
    \end{tabular}
    \caption{Comparisons of Temporal Action Grounding task on temporal test dataset. 
    MoChat-R refers to MoChat with a regression head. 
    The $\uparrow$ symbol indicates that a higher value is better. 
    Bold and underline indicate the best and the second best result.}
    \label{time}
\end{table}

\subsection{Ablation Study}
\begin{figure}[htbp]
    \centering
    \includegraphics[width=1.0\columnwidth]{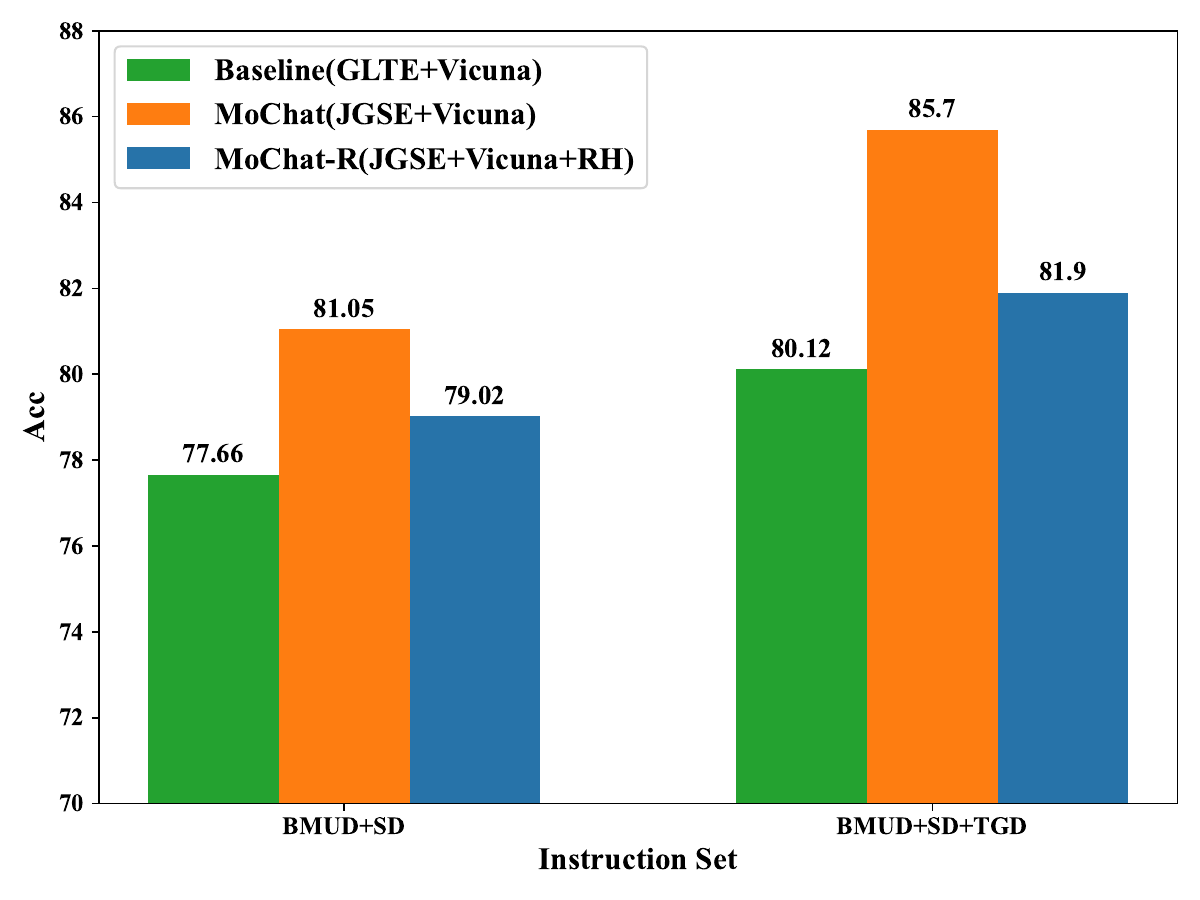} 
    \caption{Ablation study of Spatial Limb Grounding task across different models and instruction sets. 
    The module names GLTE, JGSE, and RH refer to Global-Local Transformer Encoder, Joints-Grouped Skeleton Encoder, and Regression Head, respectively.
    BMUD+SD+TGD refers to model jointly trained on Basic Motion Understanding Dialogue, Spatial Dialogue and Temporal Grounding Dialogue. 
    }
    \label{ablationAcc}    
\end{figure}

We conduct ablation studies on different combinations of instruction sets to verify the effectiveness of various components of our method. 
Specifically, we performed ablation experiments using incrementally combined instruction sets across the three tasks mentioned above. 
The results are shown in Tab. \ref{ablationMU}, Fig. \ref{ablationAcc}, and Tab. \ref{time}. As can be observed, for the same model, training with multiple instruction sets has a positive impact on the same task, demonstrating the advantages of integrated training. For the same instruction set, the model without the regression head performs best in motion understanding and spatial limb grounding tasks, while the model with the regression head performs best in the temporal action grounding task, proving the effectiveness of this module.
Additional ablation studies are included in the supplementary material.


\section{Conclusion}
In this paper, we present MoChat, a motion understanding multimodal large language model that comprehends motion sequences, 
accurately captions the movement of specific body parts, 
and precisely identifies the time boundaries corresponding to user instructions. 
To the best of our knowledge, MoChat is the first MLLM capable of spatio-temporal grounding of actions in single skeleton sequences.

Despite its promising results, MoChat has some limitations, particularly in real-time performance and resource consumption, where it does not perform as efficiently as fixed-class action recognition models. However, MoChat has significant potential for application in fields such as sports analytics, human-computer interaction, and medical rehabilitation. By advancing the ability to interpret and ground motion sequences in a spatio-temporal context, MoChat contributes to the broader development of multimodal large language models and opens up new avenues for research in motion understanding and beyond.

\bibliography{MoChat1}

\begin{thebibliography}{39}
\providecommand{\natexlab}[1]{#1}

\bibitem[{Alayrac et~al.(2024)Alayrac, Donahue, Luc, Miech, Barr, Hasson, Lenc, Mensch, Millicah, Reynolds, Ring, Rutherford, Cabi, Han, Gong, Samangooei, Monteiro, Menick, Borgeaud, Brock, Nematzadeh, Sharifzadeh, Binkowski, Barreira, Vinyals, Zisserman, and Simonyan}]{Flamingo}
Alayrac, J.-B.; Donahue, J.; Luc, P.; Miech, A.; Barr, I.; Hasson, Y.; Lenc, K.; Mensch, A.; Millicah, K.; Reynolds, M.; Ring, R.; Rutherford, E.; Cabi, S.; Han, T.; Gong, Z.; Samangooei, S.; Monteiro, M.; Menick, J.; Borgeaud, S.; Brock, A.; Nematzadeh, A.; Sharifzadeh, S.; Binkowski, M.; Barreira, R.; Vinyals, O.; Zisserman, A.; and Simonyan, K. 2024.
\newblock Flamingo: a visual language model for few-shot learning.
\newblock In \emph{Proceedings of the 36th International Conference on Neural Information Processing Systems}, NIPS '22. Red Hook, NY, USA: Curran Associates Inc.
\newblock ISBN 9781713871088.

\bibitem[{Chen et~al.(2021)Chen, Zhang, Yuan, Li, Deng, and Hu}]{ctrgcn}
Chen, Y.; Zhang, Z.; Yuan, C.; Li, B.; Deng, Y.; and Hu, W. 2021.
\newblock Channel-Wise {{Topology Refinement Graph Convolution}} for {{Skeleton-Based Action Recognition}}.
\newblock In \emph{2021 {{IEEE}}/{{CVF International Conference}} on {{Computer Vision}} ({{ICCV}})}, 13339--13348.

\bibitem[{Chen et~al.(2022)Chen, Zhao, Yuan, Tian, Xia, Geng, Han, and Metaxas}]{hitrs}
Chen, Y.; Zhao, L.; Yuan, J.; Tian, Y.; Xia, Z.; Geng, S.; Han, L.; and Metaxas, D.~N. 2022.
\newblock Hierarchically {{Self-supervised Transformer}} for~{{Human Skeleton Representation Learning}}.
\newblock In Avidan, S.; Brostow, G.; Ciss{\'e}, M.; Farinella, G.~M.; and Hassner, T., eds., \emph{Computer {{Vision}} -- {{ECCV}} 2022}, Lecture {{Notes}} in {{Computer Science}}, 185--202. Cham: Springer Nature Switzerland.
\newblock ISBN 978-3-031-19809-0.

\bibitem[{Chiang et~al.(2023)Chiang, Li, Lin, Sheng, Wu, Zhang, Zheng, Zhuang, Zhuang, Gonzalez, Stoica, and Xing}]{vicuna2023}
Chiang, W.-L.; Li, Z.; Lin, Z.; Sheng, Y.; Wu, Z.; Zhang, H.; Zheng, L.; Zhuang, S.; Zhuang, Y.; Gonzalez, J.~E.; Stoica, I.; and Xing, E.~P. 2023.
\newblock Vicuna: An Open-Source Chatbot Impressing GPT-4 with 90\%* ChatGPT Quality.

\bibitem[{Dai et~al.(2023)Dai, Li, Li, Tiong, Zhao, Wang, Li, Fung, and Hoi}]{dai2023instructblip}
Dai, W.; Li, J.; Li, D.; Tiong, A.; Zhao, J.; Wang, W.; Li, B.; Fung, P.; and Hoi, S. 2023.
\newblock Instruct{BLIP}: Towards General-purpose Vision-Language Models with Instruction Tuning.
\newblock In \emph{Thirty-seventh Conference on Neural Information Processing Systems}.

\bibitem[{Devlin et~al.(2019)Devlin, Chang, Lee, and Toutanova}]{devlin2018bert}
Devlin, J.; Chang, M.-W.; Lee, K.; and Toutanova, K. 2019.
\newblock {BERT}: Pre-training of Deep Bidirectional Transformers for Language Understanding.
\newblock In Burstein, J.; Doran, C.; and Solorio, T., eds., \emph{Proceedings of the 2019 Conference of the North {A}merican Chapter of the Association for Computational Linguistics: Human Language Technologies, Volume 1 (Long and Short Papers)}, 4171--4186. Minneapolis, Minnesota: Association for Computational Linguistics.

\bibitem[{Duan et~al.(2022)Duan, Zhao, Chen, Lin, and Dai}]{poseconv}
Duan, H.; Zhao, Y.; Chen, K.; Lin, D.; and Dai, B. 2022.
\newblock Revisiting {{Skeleton-Based Action Recognition}}.
\newblock In \emph{Proceedings of the {{IEEE}}/{{CVF Conference}} on {{Computer Vision}} and {{Pattern Recognition}}}, 2969--2978.

\bibitem[{GLM et~al.(2024)GLM, Zeng, Xu, Wang, Zhang, Yin, Rojas, Feng, Zhao, Lai, Yu, Wang, Sun, Zhang, Cheng, Gui, Tang, Zhang, Li, Zhao, Wu, Zhong, Liu, Huang, Zhang, Zheng, Lu, Duan, Zhang, Cao, Yang, Tam, Zhao, Liu, Xia, Zhang, Gu, Lv, Liu, Liu, Yang, Song, Zhang, An, Xu, Niu, Yang, Li, Bai, Dong, Qi, Wang, Yang, Du, Hou, and Wang}]{glm2024chatglm}
GLM, T.; Zeng, A.; Xu, B.; Wang, B.; Zhang, C.; Yin, D.; Rojas, D.; Feng, G.; Zhao, H.; Lai, H.; Yu, H.; Wang, H.; Sun, J.; Zhang, J.; Cheng, J.; Gui, J.; Tang, J.; Zhang, J.; Li, J.; Zhao, L.; Wu, L.; Zhong, L.; Liu, M.; Huang, M.; Zhang, P.; Zheng, Q.; Lu, R.; Duan, S.; Zhang, S.; Cao, S.; Yang, S.; Tam, W.~L.; Zhao, W.; Liu, X.; Xia, X.; Zhang, X.; Gu, X.; Lv, X.; Liu, X.; Liu, X.; Yang, X.; Song, X.; Zhang, X.; An, Y.; Xu, Y.; Niu, Y.; Yang, Y.; Li, Y.; Bai, Y.; Dong, Y.; Qi, Z.; Wang, Z.; Yang, Z.; Du, Z.; Hou, Z.; and Wang, Z. 2024.
\newblock ChatGLM: A Family of Large Language Models from GLM-130B to GLM-4 All Tools.
\newblock arXiv:2406.12793.

\bibitem[{Guo et~al.(2022{\natexlab{a}})Guo, Zou, Zuo, Wang, Ji, Li, and Cheng}]{humanml3d}
Guo, C.; Zou, S.; Zuo, X.; Wang, S.; Ji, W.; Li, X.; and Cheng, L. 2022{\natexlab{a}}.
\newblock Generating Diverse and Natural 3D Human Motions From Text.
\newblock In \emph{Proceedings of the IEEE/CVF Conference on Computer Vision and Pattern Recognition (CVPR)}, 5152--5161.

\bibitem[{Guo et~al.(2022{\natexlab{b}})Guo, Zuo, Wang, and Cheng}]{tm2t}
Guo, C.; Zuo, X.; Wang, S.; and Cheng, L. 2022{\natexlab{b}}.
\newblock TM2T: Stochastic and Tokenized Modeling for the Reciprocal Generation of 3D Human Motions and Texts.
\newblock In Avidan, S.; Brostow, G.; Ciss{\'e}, M.; Farinella, G.~M.; and Hassner, T., eds., \emph{Computer Vision -- ECCV 2022}, 580--597. Cham: Springer Nature Switzerland.
\newblock ISBN 978-3-031-19833-5.

\bibitem[{Guo et~al.(2022{\natexlab{c}})Guo, Liu, Chen, Liu, Wang, and Ding}]{crossclr}
Guo, T.; Liu, H.; Chen, Z.; Liu, M.; Wang, T.; and Ding, R. 2022{\natexlab{c}}.
\newblock Contrastive {{Learning}} from {{Extremely Augmented Skeleton Sequences}} for {{Self-Supervised Action Recognition}}.
\newblock \emph{Proceedings of the AAAI Conference on Artificial Intelligence}, 36(1): 762--770.

\bibitem[{Hong et~al.(2024)Hong, Wang, Lv, Xu, Yu, Ji, Wang, Wang, Dong, Ding, and Tang}]{Hong2024CVPR}
Hong, W.; Wang, W.; Lv, Q.; Xu, J.; Yu, W.; Ji, J.; Wang, Y.; Wang, Z.; Dong, Y.; Ding, M.; and Tang, J. 2024.
\newblock CogAgent: A Visual Language Model for GUI Agents.
\newblock In \emph{Proceedings of the IEEE/CVF Conference on Computer Vision and Pattern Recognition (CVPR)}, 14281--14290.

\bibitem[{Huang et~al.(2020)Huang, Huang, Ouyang, and Wang}]{Huang_Huang_Ouyang_Wang_2020}
Huang, L.; Huang, Y.; Ouyang, W.; and Wang, L. 2020.
\newblock Part-Level Graph Convolutional Network for Skeleton-Based Action Recognition.
\newblock \emph{Proceedings of the AAAI Conference on Artificial Intelligence}, 34(07): 11045--11052.

\bibitem[{Jiang et~al.(2024)Jiang, Chen, Liu, Yu, Yu, and Chen}]{jiang2024motiongpt}
Jiang, B.; Chen, X.; Liu, W.; Yu, J.; Yu, G.; and Chen, T. 2024.
\newblock Motiongpt: Human motion as a foreign language.
\newblock \emph{Advances in Neural Information Processing Systems}, 36.

\bibitem[{Jin et~al.(2024)Jin, Takanobu, Zhang, Cao, and Yuan}]{chatunivi}
Jin, P.; Takanobu, R.; Zhang, W.; Cao, X.; and Yuan, L. 2024.
\newblock Chat-UniVi: Unified Visual Representation Empowers Large Language Models with Image and Video Understanding.
\newblock In \emph{Proceedings of the IEEE/CVF Conference on Computer Vision and Pattern Recognition (CVPR)}, 13700--13710.

\bibitem[{Kim et~al.(2022)Kim, Chang, Kim, and Choi}]{gltrans}
Kim, B.; Chang, H.~J.; Kim, J.; and Choi, J.~Y. 2022.
\newblock Global-Local Motion Transformer for Unsupervised Skeleton-Based Action Learning.
\newblock In Avidan, S.; Brostow, G.; Ciss{\'e}, M.; Farinella, G.~M.; and Hassner, T., eds., \emph{Computer Vision -- ECCV 2022}, 209--225. Cham: Springer Nature Switzerland.
\newblock ISBN 978-3-031-19772-7.

\bibitem[{Li et~al.(2023)Li, Li, Savarese, and Hoi}]{blip2}
Li, J.; Li, D.; Savarese, S.; and Hoi, S. 2023.
\newblock {BLIP}-2: Bootstrapping Language-Image Pre-training with Frozen Image Encoders and Large Language Models.
\newblock In Krause, A.; Brunskill, E.; Cho, K.; Engelhardt, B.; Sabato, S.; and Scarlett, J., eds., \emph{Proceedings of the 40th International Conference on Machine Learning}, volume 202 of \emph{Proceedings of Machine Learning Research}, 19730--19742. PMLR.

\bibitem[{Li et~al.(2024)Li, He, Wang, Li, Wang, Luo, Wang, Wang, and Qiao}]{2023videochat}
Li, K.; He, Y.; Wang, Y.; Li, Y.; Wang, W.; Luo, P.; Wang, Y.; Wang, L.; and Qiao, Y. 2024.
\newblock VideoChat: Chat-Centric Video Understanding.
\newblock arXiv:2305.06355.

\bibitem[{Lin(2004)}]{lin2004rouge}
Lin, C.-Y. 2004.
\newblock {ROUGE}: A Package for Automatic Evaluation of Summaries.
\newblock In \emph{Text Summarization Branches Out}, 74--81. Barcelona, Spain: Association for Computational Linguistics.

\bibitem[{Liu et~al.(2024)Liu, Li, Li, and Lee}]{llava1.5}
Liu, H.; Li, C.; Li, Y.; and Lee, Y.~J. 2024.
\newblock Improved Baselines with Visual Instruction Tuning.
\newblock In \emph{Proceedings of the IEEE/CVF Conference on Computer Vision and Pattern Recognition (CVPR)}, 26296--26306.

\bibitem[{Loper et~al.(2015)Loper, Mahmood, Romero, Pons-Moll, and Black}]{SMPL:2015}
Loper, M.; Mahmood, N.; Romero, J.; Pons-Moll, G.; and Black, M.~J. 2015.
\newblock {SMPL}: A Skinned Multi-Person Linear Model.
\newblock \emph{ACM Trans. Graphics (Proc. SIGGRAPH Asia)}, 34(6): 248:1--248:16.

\bibitem[{OpenAI(2024)}]{openai2024gpt4technicalreport}
OpenAI. 2024.
\newblock GPT-4 Technical Report.
\newblock arXiv:2303.08774.

\bibitem[{Papineni et~al.(2002)Papineni, Roukos, Ward, and Zhu}]{papineni2002bleu}
Papineni, K.; Roukos, S.; Ward, T.; and Zhu, W.-J. 2002.
\newblock BLEU: a method for automatic evaluation of machine translation.
\newblock In \emph{Proceedings of the 40th Annual Meeting on Association for Computational Linguistics}, ACL '02, 311–318. USA: Association for Computational Linguistics.

\bibitem[{Qian et~al.(2024)Qian, Li, Wu, Ye, Fei, Chua, Zhuang, and Tang}]{momentor}
Qian, L.; Li, J.; Wu, Y.; Ye, Y.; Fei, H.; Chua, T.-S.; Zhuang, Y.; and Tang, S. 2024.
\newblock Momentor: Advancing Video Large Language Model with Fine-Grained Temporal Reasoning.
\newblock In Salakhutdinov, R.; Kolter, Z.; Heller, K.; Weller, A.; Oliver, N.; Scarlett, J.; and Berkenkamp, F., eds., \emph{Proceedings of the 41st International Conference on Machine Learning}, volume 235 of \emph{Proceedings of Machine Learning Research}, 41340--41356. PMLR.

\bibitem[{Radford et~al.(2021)Radford, Kim, Hallacy, Ramesh, Goh, Agarwal, Sastry, Askell, Mishkin, Clark, Krueger, and Sutskever}]{CLIP}
Radford, A.; Kim, J.~W.; Hallacy, C.; Ramesh, A.; Goh, G.; Agarwal, S.; Sastry, G.; Askell, A.; Mishkin, P.; Clark, J.; Krueger, G.; and Sutskever, I. 2021.
\newblock Learning Transferable Visual Models From Natural Language Supervision.
\newblock In \emph{International Conference on Machine Learning}.

\bibitem[{Ren et~al.(2024)Ren, Yao, Li, Sun, and Hou}]{timechat}
Ren, S.; Yao, L.; Li, S.; Sun, X.; and Hou, L. 2024.
\newblock TimeChat: A Time-sensitive Multimodal Large Language Model for Long Video Understanding.
\newblock In \emph{Proceedings of the IEEE/CVF Conference on Computer Vision and Pattern Recognition (CVPR)}, 14313--14323.

\bibitem[{Shi et~al.(2019)Shi, Zhang, Cheng, and Lu}]{2sagcn}
Shi, L.; Zhang, Y.; Cheng, J.; and Lu, H. 2019.
\newblock Two-{{Stream Adaptive Graph Convolutional Networks}} for {{Skeleton-Based Action Recognition}}.
\newblock In \emph{2019 {{IEEE}}/{{CVF Conference}} on {{Computer Vision}} and {{Pattern Recognition}} ({{CVPR}})}, 12018--12027. Long Beach, CA, USA: IEEE.
\newblock ISBN 978-1-72813-293-8.

\bibitem[{Van Den~Oord, Vinyals et~al.(2017)}]{vqvae}
Van Den~Oord, A.; Vinyals, O.; et~al. 2017.
\newblock Neural discrete representation learning.
\newblock \emph{Advances in neural information processing systems}, 30.

\bibitem[{Vaswani et~al.(2017)Vaswani, Shazeer, Parmar, Uszkoreit, Jones, Gomez, Kaiser, and Polosukhin}]{transformer}
Vaswani, A.; Shazeer, N.; Parmar, N.; Uszkoreit, J.; Jones, L.; Gomez, A.~N.; Kaiser, L.~u.; and Polosukhin, I. 2017.
\newblock Attention is All you Need.
\newblock In Guyon, I.; Luxburg, U.~V.; Bengio, S.; Wallach, H.; Fergus, R.; Vishwanathan, S.; and Garnett, R., eds., \emph{Advances in Neural Information Processing Systems}, volume~30. Curran Associates, Inc.

\bibitem[{Vedantam, Lawrence~Zitnick, and Parikh(2015)}]{vedantam2015cider}
Vedantam, R.; Lawrence~Zitnick, C.; and Parikh, D. 2015.
\newblock CIDEr: Consensus-Based Image Description Evaluation.
\newblock In \emph{Proceedings of the IEEE Conference on Computer Vision and Pattern Recognition (CVPR)}.

\bibitem[{Wang, Zhang, and Asghar(2022)}]{stgcn}
Wang, Q.; Zhang, K.; and Asghar, M.~A. 2022.
\newblock Skeleton-Based ST-GCN for Human Action Recognition With Extended Skeleton Graph and Partitioning Strategy.
\newblock \emph{IEEE Access}, 10: 41403--41410.

\bibitem[{Wang et~al.(2024)Wang, Lv, Yu, Hong, Qi, Wang, Ji, Yang, Zhao, Song, Xu, Xu, Li, Dong, Ding, and Tang}]{wang2024cogvlmvisualexpertpretrained}
Wang, W.; Lv, Q.; Yu, W.; Hong, W.; Qi, J.; Wang, Y.; Ji, J.; Yang, Z.; Zhao, L.; Song, X.; Xu, J.; Xu, B.; Li, J.; Dong, Y.; Ding, M.; and Tang, J. 2024.
\newblock CogVLM: Visual Expert for Pretrained Language Models.
\newblock arXiv:2311.03079.

\bibitem[{Yan et~al.(2023)Yan, Liu, Wei, Li, and Lin}]{yan2023skeletonmae}
Yan, H.; Liu, Y.; Wei, Y.; Li, G.; and Lin, L. 2023.
\newblock SkeletonMAE: Graph-based Masked Autoencoder for Skeleton Sequence Pre-training.
\newblock In \emph{Proceedings of the IEEE/CVF International Conference on Computer Vision}.

\bibitem[{Zhang et~al.(2024)Zhang, Herrmann, Hur, Chen, Jampani, Sun, and Yang}]{TellingLeft}
Zhang, J.; Herrmann, C.; Hur, J.; Chen, E.; Jampani, V.; Sun, D.; and Yang, M.-H. 2024.
\newblock Telling Left from Right: Identifying Geometry-Aware Semantic Correspondence.
\newblock In \emph{Proceedings of the IEEE/CVF Conference on Computer Vision and Pattern Recognition (CVPR)}, 3076--3085.

\bibitem[{Zhang* et~al.(2020)Zhang*, Kishore*, Wu*, Weinberger, and Artzi}]{BERTScore}
Zhang*, T.; Kishore*, V.; Wu*, F.; Weinberger, K.~Q.; and Artzi, Y. 2020.
\newblock BERTScore: Evaluating Text Generation with BERT.
\newblock In \emph{International Conference on Learning Representations}.

\bibitem[{Zheng et~al.(2023)Zheng, Chiang, Sheng, Zhuang, Wu, Zhuang, Lin, Li, Li, Xing, Zhang, Gonzalez, and Stoica}]{zheng2023judging}
Zheng, L.; Chiang, W.-L.; Sheng, Y.; Zhuang, S.; Wu, Z.; Zhuang, Y.; Lin, Z.; Li, Z.; Li, D.; Xing, E.; Zhang, H.; Gonzalez, J.~E.; and Stoica, I. 2023.
\newblock Judging {LLM}-as-a-Judge with {MT}-Bench and Chatbot Arena.
\newblock In \emph{Thirty-seventh Conference on Neural Information Processing Systems Datasets and Benchmarks Track}.

\bibitem[{Zheng et~al.(2020)Zheng, Wang, Liu, Li, Ye, and Ren}]{zheng2020diou}
Zheng, Z.; Wang, P.; Liu, W.; Li, J.; Ye, R.; and Ren, D. 2020.
\newblock Distance-IoU Loss: Faster and Better Learning for Bounding Box Regression.
\newblock In \emph{The AAAI Conference on Artificial Intelligence (AAAI)}, 12993--13000.

\bibitem[{Zhou, Wan, and Wang(2024)}]{AvatarGPT}
Zhou, Z.; Wan, Y.; and Wang, B. 2024.
\newblock AvatarGPT: All-in-One Framework for Motion Understanding Planning Generation and Beyond.
\newblock In \emph{Proceedings of the IEEE/CVF Conference on Computer Vision and Pattern Recognition (CVPR)}, 1357--1366.

\bibitem[{Zhu et~al.(2024)Zhu, Lin, Ning, Yan, Cui, HongFa, Pang, Jiang, Zhang, Li, Zhang, Li, Liu, and Yuan}]{zhu2024languagebind}
Zhu, B.; Lin, B.; Ning, M.; Yan, Y.; Cui, J.; HongFa, W.; Pang, Y.; Jiang, W.; Zhang, J.; Li, Z.; Zhang, C.~W.; Li, Z.; Liu, W.; and Yuan, L. 2024.
\newblock LanguageBind: Extending Video-Language Pretraining to N-modality by Language-based Semantic Alignment.
\newblock In \emph{The Twelfth International Conference on Learning Representations}.

\end{thebibliography}

\section{Data Construction}
\begin{figure}[htbp]
    \centering
    \includegraphics[width=1.0\columnwidth]{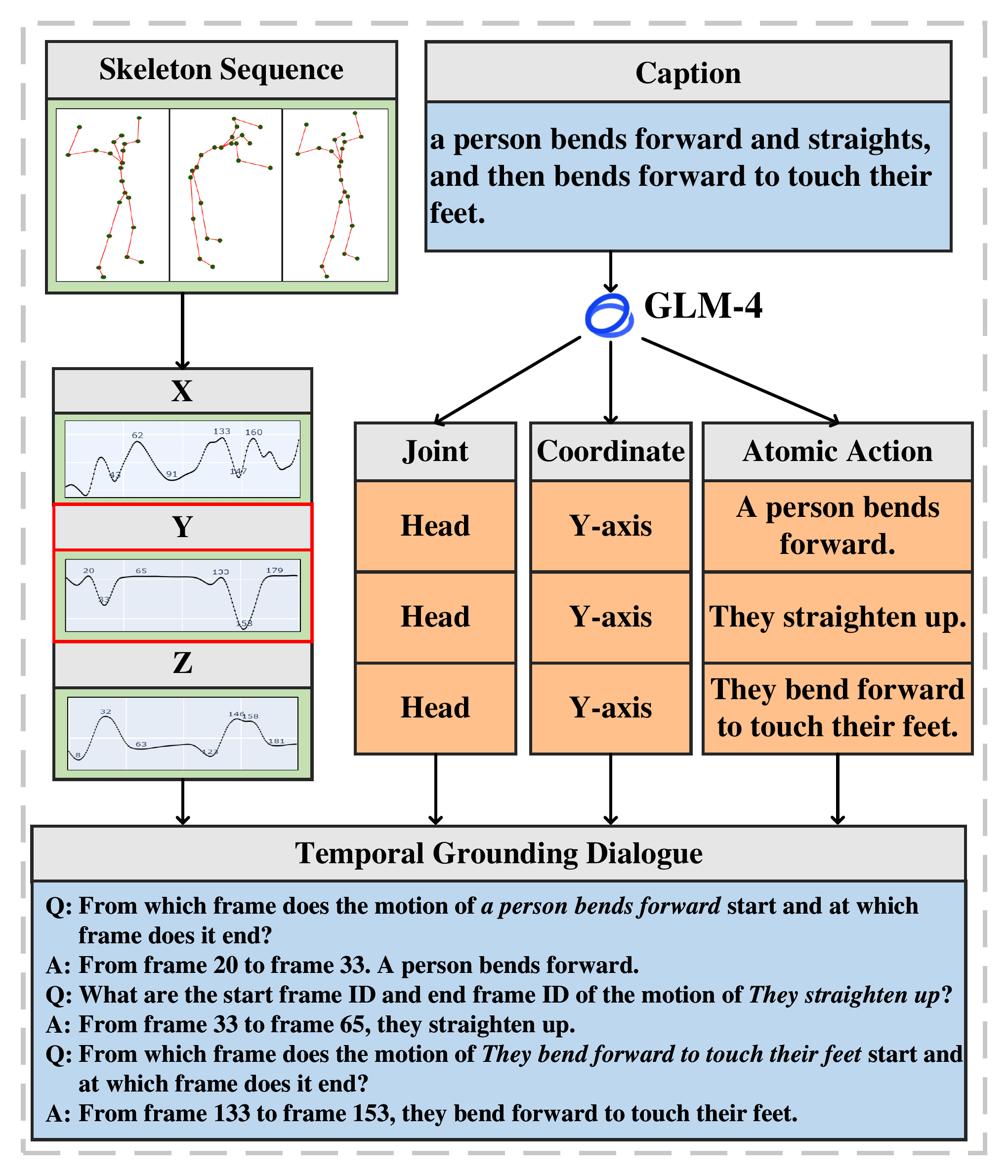} 
	\caption{Pipeline for constructing Temporal Grounding Dialogues. 
	GLM-4 splits the caption into atomic actions and identifies the corresponding most significant joint and coordinate. 
	The curves represent the coordinates of the selected joint, 
	with the numbers on the curves indicating the frame IDs of the extremum points. 
	We construct multi-turn temporal grounding dialogues based on the final extracted results.}
	
    \label{pipeline}
\end{figure}

\begin{figure*}
    \centering
    \includegraphics[width=1.0\linewidth]{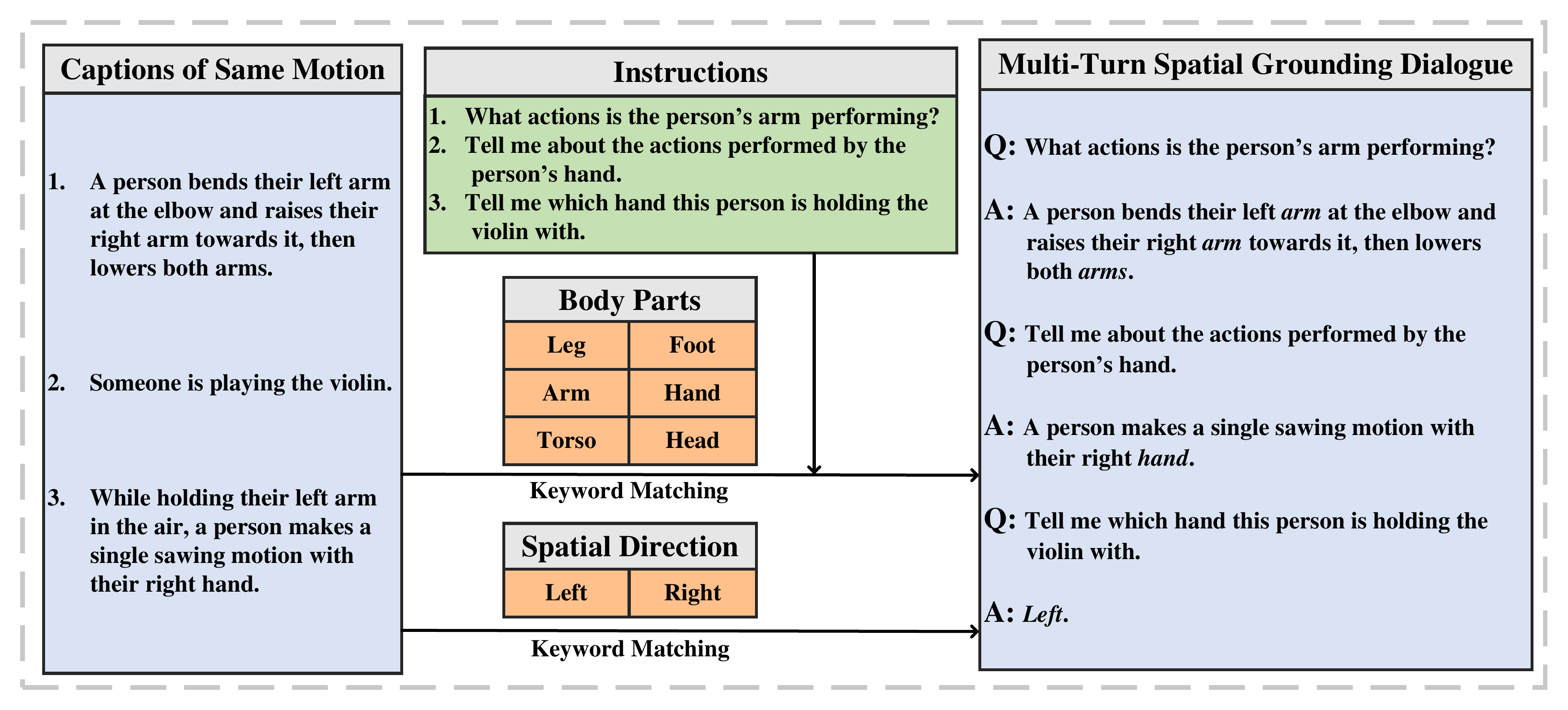}
    \caption{The process of constructing Spatial Dialogues.}
    \label{spatial_dialogues}
\end{figure*}

\begin{figure*}[htbp]
    \centering
    \includegraphics[width=2.0\columnwidth]{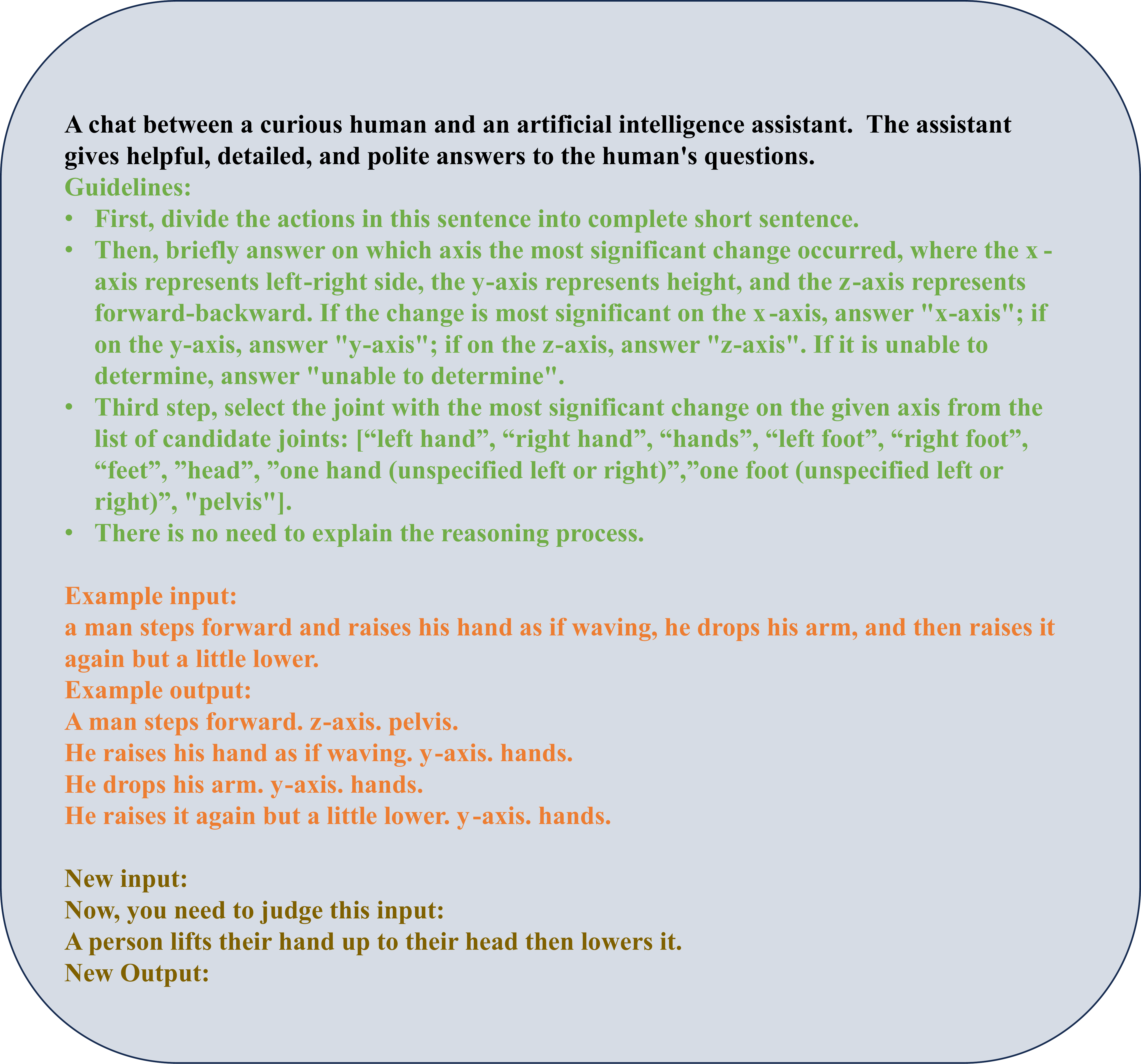} 
    \caption{Instructions provided to GLM-4 for splitting captions.}
    \label{glm4instruction}
\end{figure*}

Tab. \ref{full_instructions} presents all the templates used to construct the dialogues. 
As previously described, for each task, we randomly select an instruction from the instruction set 
and then generate the corresponding response based on the dataset. 
The process for constructing the Spatial Dialogue is illustrated in Fig. \ref{spatial_dialogues}. 
We perform keyword matching on the captions, with each keyword generating a dialogue turn, ultimately forming a multi-turn dialogue. 
The pipeline for constructing the Temporal Grounding Dialogue is illustrated in Fig. \ref{pipeline}. 
In this process, we use GLM-4 to process the captions and apply the instruction shown in Fig. \ref{glm4instruction} to filter the joints and axes with the most significant variation. 
Finally, the motion data is utilized to determine the start and end frame IDs.

\section{Implementation Details}
Aside from the aforementioned use of Vicuna-v1.5-13B as the language foundation model, 
all our models employ the AdamW optimizer for training. 
For the skeleton encoder pre-training, 
we use a batch size of 128 and train the model for 120 epochs with a learning rate of \(5 \times 10^{-5}\) 
and a decay rate of 0.99. 
The encoder consists of a 4-layer transformer. The input sequences are padded to 500 frames with a value of 99.9. 
To align the motion-language embeddings, the model is trained with a batch size of 64 for 3 epochs, 
using a learning rate of \(2 \times 10^{-3}\). 
The learning rate schedule includes a warm-up ratio of 0.03, 
followed by cosine annealing. In the final stage, for fine-tuning the model end-to-end, a batch size of 128 is applied, 
with training conducted over 1 epoch at a learning rate of \(2 \times 10^{-5}\). 
The same warm-up and cosine annealing schedule from the previous stage is utilized. 
When GPU memory is insufficient, 
we reduce the per\_device\_train\_batch\_size and increase the \text{gradient\_accumulation\_steps} while keeping the product of \text{per\_device\_train\_batch\_size}, \text{GPU\_num}, 
and \text{gradient\_accumulation\_steps} equal to the original batch size. 
The training duration is approximately 8 hours for the skeleton encoder pre-training, 
10 hours for aligning the motion-language embeddings, and 5 hours for the final fine-tuning.

\section{Additional Experiments}
\subsubsection{LoRA Parameters and Model Size}
We explore solutions to reduce resource consumption by experimenting with LoRA and different language foundation model sizes. 
As shown in Tab. \ref{additionalMU}, we train and evaluate the model with a LoRA rank of 64 and an alpha of 16, 
and separately experiment with a language foundation model with 7B parameters. 
However, compared to the 13B model, while the memory usage is reduced, 
the resulting performance degradation was unacceptable. 
This indicates the need to explore other, more effective methods for reducing memory consumption.

\subsubsection{Custom FrameID Tokens}
In addition to the Regression Head, 
we also experiment with using custom frame ID tokens (CFT) to identify the start and end frames corresponding to the captions. 
Specifically, We add \(T\) tokens to the tokenizer's vocabulary, 
such as \texttt{<frameid\_0>}, \texttt{<frameid\_1>}, ..., \texttt{<frameid\_T>}. 
Similar to positional encoding, we obtain their corresponding embeddings 
and add them to the motion token embeddings, before finally inserting them into the language embeddings. 

As shown by the metrics across the three tasks (Tab. \ref{additionalLR}, \ref{additionalTime} and \ref{additionalMU}), 
while the use of CFT results in better performance on the Motion Understanding and Spatial Limb Grounding tasks, 
it still underperforms compared to the Regression Head on the Temporal Action Grounding task.

\subsubsection{Instruction Set Configuration}
During the fine-tuning of the LLM, we observe catastrophic forgetting, 
where the model lost its ability to follow general instructions, a capability typically possessed by the base model. 
To preserve the model's instruction-following ability, we utilize the Puffin dataset, a subset of processed ShareGPT data, 
containing 3,000 examples, with each response generated using GPT-4. 
As shown in Tab. \ref{additionalMU}, the results indicate that, without using the Puffin dataset, 
some metrics for the motion understanding task improve. 
However, the model fails to generate reasonable responses to other types of instructions, 
such as “Who are you?”—a question unrelated to the motion understanding task—resulting in a less user-friendly model.

Additionally, we explore the impact of using different instruction sets at various stages of instruction fine-tuning. 
For instance, we use Basic Motion Understanding Dialogues during the alignment of motion-language embeddings, 
and combine Basic Motion Understanding Dialogues, Spatial Dialogues, and Temporal Grounding Dialogues during full fine-tuning. 
As shown by the results in the tables (Tab. \ref{additionalLR}, \ref{additionalTime} and \ref{additionalMU}), when the same instruction set is used across both stages, 
the model performs better on the Motion Understanding and Spatial Limb Grounding tasks, but worse on the Temporal Action Grounding task.

\section{Analysis of Learned Attention}
\begin{figure*}[htbp]
    \centering
    \includegraphics[width=1.0\textwidth]{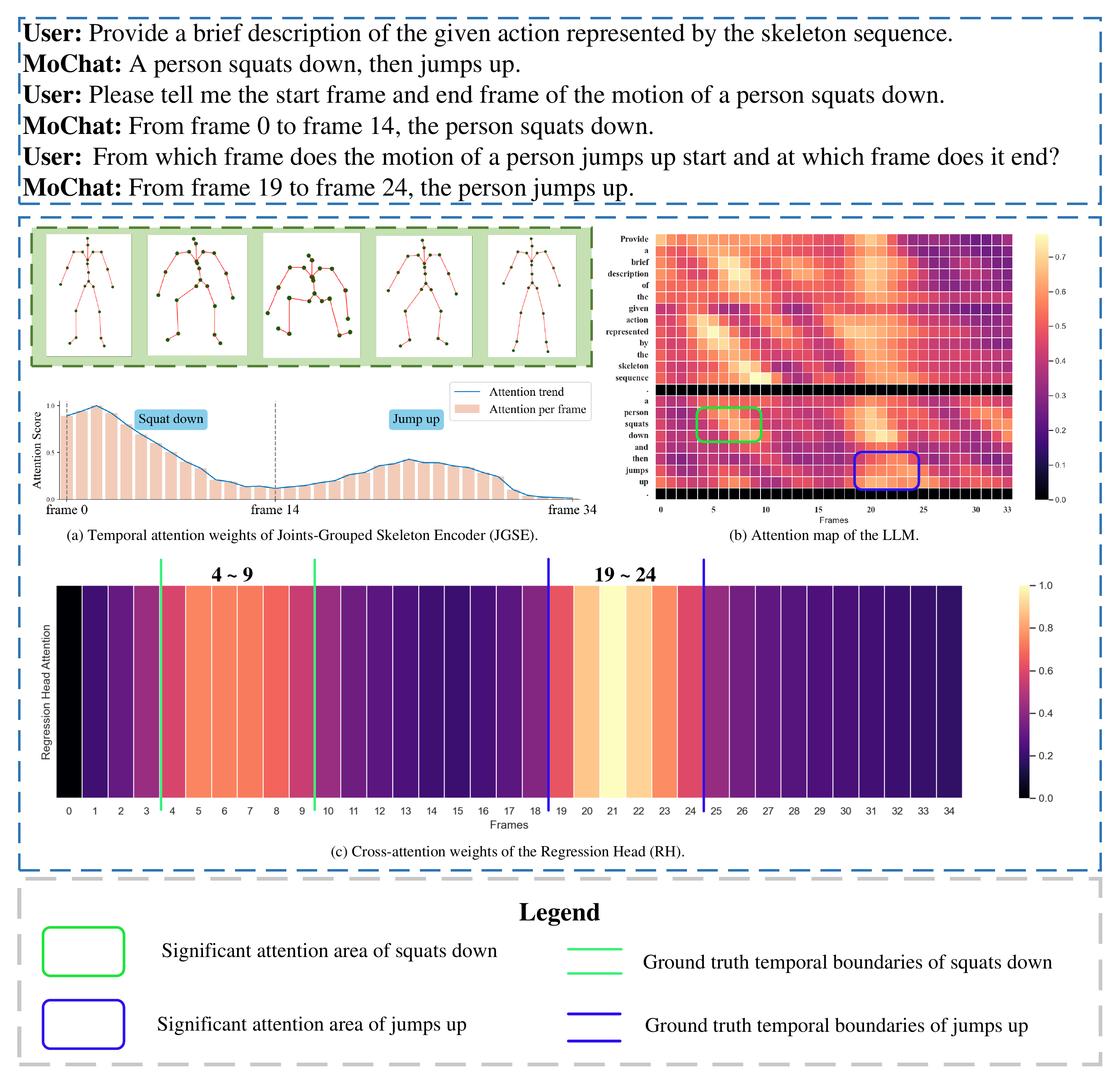} 
    \caption{Attention visualization of three modules.}
    \label{attention}
\end{figure*}

To gain further insights into our model, we visualize the attention weights of the Joints-Grouped Skeleton Encoder (JGSE), the LLM, and the Regression Head modules. 
For the JGSE, we compute the average self-attention weights from the last layer of the Transformer Encoder 
and then visualize the attention of the last temporal \texttt{[CLS]} token to other skeleton frames, as shown in Fig. \ref{attention} (a). 
We concatenate the resulting motion embeddings with the language embeddings and feed them into the LLM, 
then extract the attention matrix from the first head of the first layer. 
The attention weights are averaged across multiple language tokens to form complete words, as depicted in Fig. \ref{attention} (b). 
For the Regression Head, we visualize the cross-attention weights of the \texttt{[BOS]} token with respect to the motion embeddings, 
as shown in Fig. \ref{attention} (c). 
The attention visualizations from these three modules demonstrate that our model effectively captures temporal awareness 
and motion-caption mapping, enabling it to successfully perform the Temporal Action Grounding task.

\setlength{\tabcolsep}{6pt}
\begin{table}[htbp]
	\begin{tabular}{c|c|c|c}
	\toprule
	\textbf{Module}   & \textbf{Stage1}  & \textbf{Stage2}  & \textbf{Acc.} \\
	\midrule
	GLTE+Vicuna-13B            & \multicolumn{2}{c|}{\multirow{2}{*}{BS}}     & 77.66          \\
	GLTE+Vicuna-7B             & \multicolumn{2}{c|}{}                             & 73.10          \\
	\midrule
	JGSE+CFT+Vicuna-13B        & B                 & BST              & 85.28          \\
	\midrule
	JGSE+CFT+Vicuna-13B        & \multicolumn{2}{c|}{\multirow{3}{*}{BST}} & \textbf{85.79} \\
	JGSE+Vicuna-13B            & \multicolumn{2}{c|}{}                             & 85.70   \\
	JGSE+RH+Vicuna-13B         & \multicolumn{2}{c|}{}                             & 81.90          \\
	\bottomrule
	\end{tabular}
	\caption{Additional experiments for Spatial Limb Grounding task, 
	The module names GLTE, JGSE, CFT and RH refer to Global-Local Transformer Encoder, Joints-Grouped Skeleton Encoder, Custom FrameID Tokens and Regression Head, respectively. 
    BST indicates that the model was jointly trained on Basic Motion Understanding Dialogue, Spatial Dialogue, and Temporal Grounding Dialogue. 
    A higher value is better. Bold indicates the best result.}
	\label{additionalLR}
\end{table}

\setlength{\tabcolsep}{1pt}
\begin{table}[htbp]
	\begin{tabular}{c|c|c|cc}
	\toprule
	\textbf{Module}   & \textbf{Stage1}  & \textbf{Stage2}  & \textbf{R@1(IoU=0.5)} & \textbf{R@1(IoU=0.7)} \\
	\midrule
	JGSE+CFT & B                 & BST              & 20.17             & 9.01              \\
	\midrule
	JGSE+CFT & \multicolumn{2}{c|}{\multirow{2}{*}{BST}} & 18.03             & 7.30              \\
	JGSE+RH  & \multicolumn{2}{c|}{}                             & \textbf{21.89}    & \textbf{12.02}   \\
	\bottomrule
	\end{tabular}
	\caption{Additional experiments for the Temporal Action Grounding task.
	The module names JGSE, CFT, and RH refer to Joints-Grouped Skeleton Encoder, Custom FrameID Tokens, and Regression Head, respectively. 
	BST indicates that the model was jointly trained on Basic Motion Understanding Dialogue, Spatial Dialogue, and Temporal Grounding Dialogue. 
	R@1 denotes Recall at rank 1 for IoU thresholds of 0.5 and 0.7, with higher values indicating better performance. 
	Bold values indicate the best results.}
	\label{additionalTime}
\end{table}

\setlength{\tabcolsep}{2pt}
\begin{table*}[htbp]
	\centering
	\begin{tabular}{c|c|c|c|cccccc}
	\toprule
	\textbf{Module} & \textbf{Lora} & \textbf{Stage1} & \textbf{Stage2} & \textbf{BLEU@1} & \textbf{BLEU@4} & \textbf{ROUGE} & \textbf{CIDEr} & \textbf{BERTScore} & \textbf{GPT4Score} \\
	\midrule
	GLTE+Vicuna-13B & -- & \multicolumn{2}{c|}{B wo Puffin} & \textbf{62.36} & \textbf{22.51} & 47.09 & 50.35 & 44.25 & 5.53 \\ 
	\midrule
	GLTE+Vicuna-13B & -- & \multicolumn{2}{c|}{\multirow{2}{*}{B}}  & 59.85 & 20.80 & 45.46 & 44.88 & 41.63 & 5.21 \\ 
	GLTE+Vicuna-13B & r=64 alpha=16 &  \multicolumn{2}{c|}{}  & 37.42 & 7.54 & 32.01 & 21.81 & 38.46 & 5.24 \\ 
	\midrule
	GLTE+Vicuna-13B & -- & \multicolumn{2}{c|}{\multirow{2}{*}{BS}}  & 59.95 & 20.51 & \textbf{47.64} & 49.30 & 44.28 & 5.80 \\ 
	GLTE+Vicuna-7B & -- & \multicolumn{2}{c|}{}  & 47.30 & 11.20 & 38.39 & 41.80 & 30.24 & 3.24 \\ 
	\midrule
	JGSE+CFT+Vicuna-13B & -- & B & BST & 59.96 & 20.88 & 46.38 & 47.11 & 43.47 & 5.50 \\ 
	\midrule
	JGSE+CFT+Vicuna-13B & -- & \multicolumn{2}{c|}{\multirow{3}{*}{BST}}  & 61.16 & 21.49 & 46.75 & 49.27 & 44.12 & \textbf{6.05} \\ 
	JGSE+RH+Vicuna-13B & -- & \multicolumn{2}{c|}{}  & 60.06 & 21.30 & 46.08 & 46.57 & 42.56 & 5.35 \\ 
	JGSE+Vicuna-13B & -- & \multicolumn{2}{c|}{}  & 61.75 & 21.60 & 47.59 & \textbf{51.57} & \textbf{45.59} & 5.99 \\ 
	\bottomrule
	\end{tabular}
	\caption{Additional experiments for Motion Understanding task, 
	\(r\) denotes the rank of the low-rank matrices, and \(alpha\) is the scaling factor controlling the impact of the adaptation.
	The module names GLTE, JGSE, CFT and RH refer to Global-Local Transformer Encoder, Joints-Grouped Skeleton Encoder, Custom FrameID Tokens and Regression Head, respectively. 
    BST indicates that the model was jointly trained on Basic Motion Understanding Dialogue, Spatial Dialogue, and Temporal Grounding Dialogue. 
    A higher value is better. Bold indicates the best result.}
	\label{additionalMU}
\end{table*}

\begin{table*}[htbp]
    \begin{tabular}{>{\centering\arraybackslash}m{2cm}>{\centering\arraybackslash}m{5cm}>{\centering\arraybackslash}m{5cm}>{\centering\arraybackslash}m{5cm}}
    \toprule
    \textbf{Input Motion Sequences} & \includegraphics[width=0.3\textwidth]{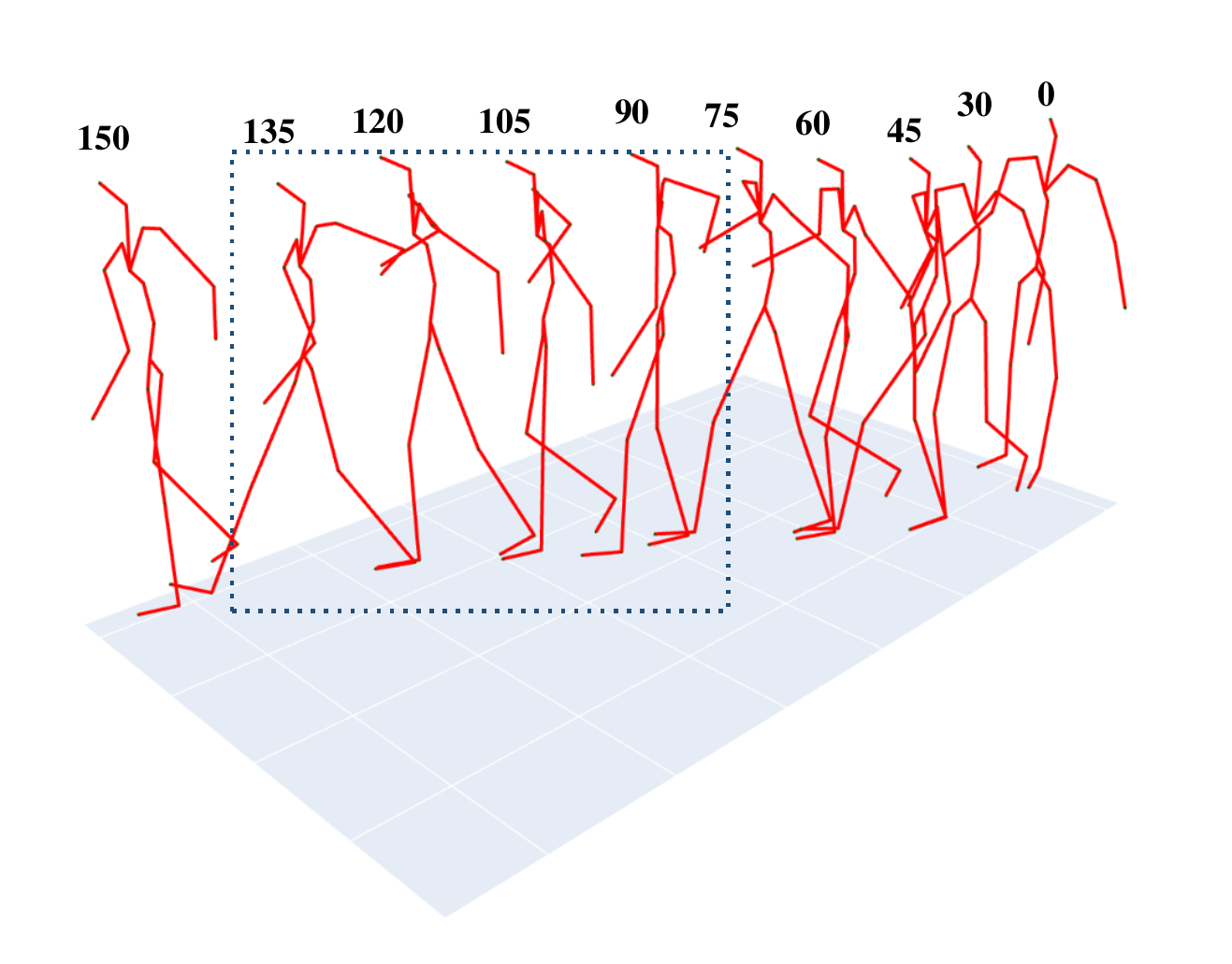} & \includegraphics[width=0.25\textwidth]{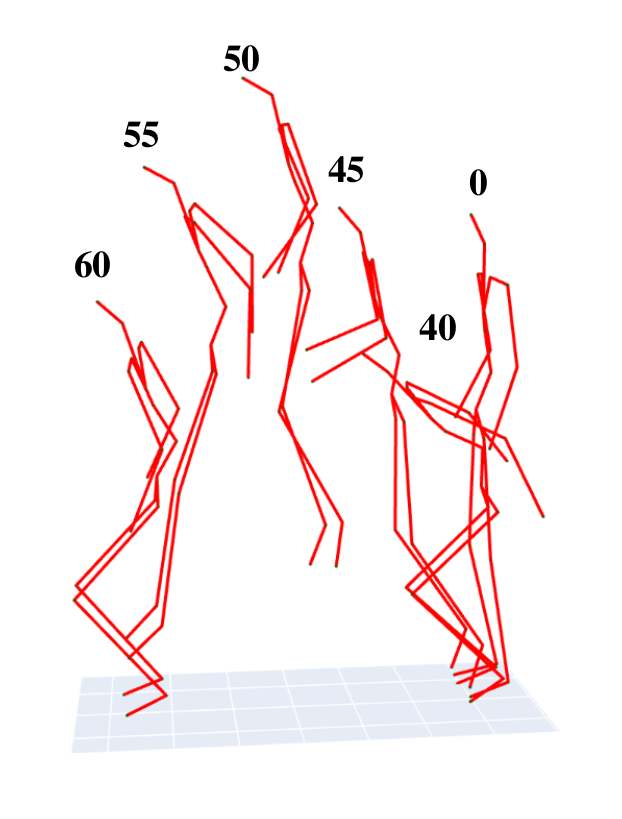} & \includegraphics[width=0.25\textwidth]{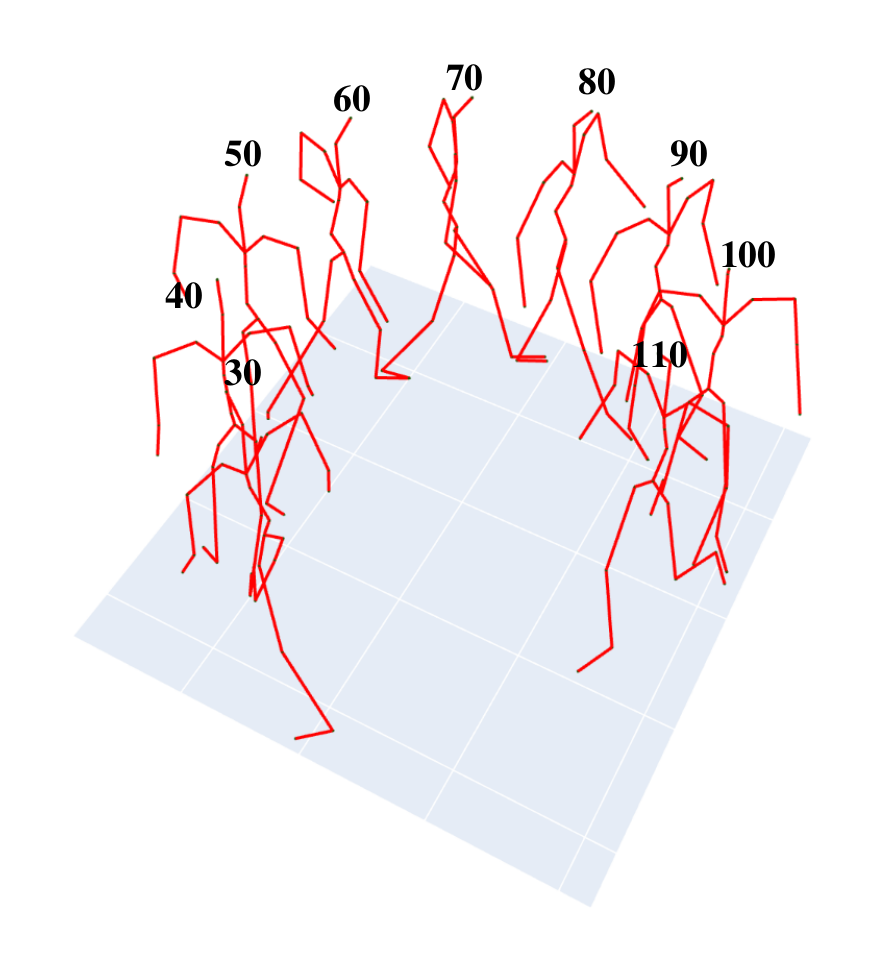} \\ 
    \midrule
    \textbf{Caption} & a person takes a step forward, moves to theor right, them continues foward with their \textbf{\textit{right hand on a rail.}} & a person jumps forward once. & a person walks in a circle, \textbf{\textit{clockwise.}} \\ 
    \textbf{MotionGPT} & a person is walking downhill. & a person jumps down a grey block. & a person walks in a circle. \\ 
    \textbf{MoChat-RH} & a person walks forward while \textbf{\textit{holding handrail with right hand.}} &a person jumps forward with both arms outstretched. & a person walks in a \textbf{\textit{ clockwise}} circle. \\ 
    \bottomrule
    \end{tabular}
    \caption{The quality results of MoChat-RH and the state-of-the-art method on the motion understanding task. 
    The results demonstrate that our method exhibits a stronger perception of action details. 
    Italics in the table indicate the matched details.}
    \label{quality_results}
\end{table*}

\begin{table*}[htbp]
    \begin{threeparttable}
        \begin{tabularx}{\textwidth}{>{\centering\arraybackslash}m{4cm}|>{\RaggedRight\arraybackslash}m{13cm}}
            \toprule
            \rowcolor{gray!10} \multicolumn{2}{c}{\textbf{Basic Motion Understanding Dialogues}} \\
            \midrule
            \textbf{Instruction Templates} &
            Provide a brief description of the given action represented by the skeleton sequence. \newline
            Write a terse but informative summary of the action depicted by the skeleton sequence. \newline
            Share a concise interpretation of the action demonstrated in the skeleton sequence. \newline
            Relay a brief, clear account of the action shown in the skeleton sequence. \newline
            Render a clear and concise summary of the action sequence. \newline
            Create a compact narrative representing the action portrayed in the skeleton sequence. \newline
            Give a short and clear explanation of the subsequent action depicted by the skeleton sequence. \newline
            Summarize the movement content of the action demonstrated by the skeleton sequence. \newline
            Describe the action concisely as represented in the skeleton sequence. \newline
            Offer a succinct explanation of the action presented in the skeleton sequence. \newline
            Present a compact description of the action sequence's key features. \\
            \midrule
            \textbf{Example} &
            \textbf{Q:} Provide a brief description of the given action represented by the skeleton sequence. \newline
            \textbf{A:} A person walks forward, then turns around and walks backward. \\
            \midrule
            \rowcolor{gray!10} \multicolumn{2}{c}{\textbf{Temporal Grounding Dialogues}} \\
            \midrule
            \textbf{Dialogue Templates} & 
            From which frame does \texttt{<motion>} start and at which frame does it end? \newline
            What are the start frame ID and end frame ID of \texttt{<motion>}? \newline
            Please tell me when \texttt{<motion>} was executed in this skeleton sequence. \newline
            From \texttt{<frameid\_1>} to \texttt{<frameid\_2>}, \texttt{<motion>}. \\
            \midrule
            \textbf{Example} & 
            \textbf{Q:} Please tell me when \textbf{\textit{A person bends forward}} was executed in this skeleton sequence. \newline
            \textbf{A:} From \textbf{\textit{frame 20}} to \textbf{\textit{frame 33}}. \textbf{\textit{A person bends forward}}. \\
            \midrule
            \rowcolor{gray!10} \multicolumn{2}{c}{\textbf{Spatial Gap-filling Dialogues}} \\
            \midrule
            \textbf{Instruction Templates} & 
            \texttt{<motion\_with\_gap>}, Complete the content in brackets with \textbf{\textit{left} or \textit{right}}. \\
            \midrule
            \textbf{Example} & 
            \textbf{Q:} Person leans forward goes onto knees whilst first putting \textbf{(\rule{5mm}{0.4pt})} hand on ground for support and stays on knees. Complete the content in brackets with left or right. \newline
            \textbf{A:} \textbf{\textit{Left}}. \\
            \midrule
            \rowcolor{gray!10} \multicolumn{2}{c}{\textbf{Spatial Multi-turn Dialogues}} \\
            \midrule
            \textbf{Instruction Templates} & 
            Describe the movements of the person’s \texttt{<body\_part>} in detail. \newline
            Please provide details about the actions of the person’s \texttt{<body\_part>}. \newline
            What actions is the person's \texttt{<body\_part>} performing?  \newline
            Tell me about the actions performed by the person's \texttt{<body\_part>}. \\
            \midrule
            \textbf{Example} & 
            \textbf{Q:} Tell me about the actions performed by the person's \textbf{\textit{torso}}. \newline
            \textbf{A:} The person walked backwards slowly, their \textbf{\textit{torso}} remaining upright, before stepping forward with a forceful kick. \newline
            \textbf{Q:} What actions is the person's \textbf{\textit{arm}} performing? \newline
            \textbf{A:} A person bends their left arm at the elbow and raises their right \textbf{\textit{arm}} towards it, then lowers both arms. \\
            \bottomrule        
        \end{tabularx}
        \caption{Dialogue Templates. \textit{Q} represents the human instruction, and \textit{A} represents the ground truth answer.}
        \label{full_instructions}
    \end{threeparttable}
\end{table*}

\end{document}